\documentclass[runningheads]{llncs}

\usepackage[mobile]{eccv}

\usepackage{eccvabbrv}

\usepackage{graphicx}
\usepackage{booktabs}

\usepackage[accsupp]{axessibility}  

\usepackage[utf8]{inputenc} 
\usepackage[T1]{fontenc}    
\usepackage{url}            
\usepackage{booktabs}       
\usepackage{amsfonts}       
\usepackage{nicefrac}       
\usepackage{microtype}      
\usepackage{xcolor}         
\usepackage{amsmath} 
\usepackage{float} 

\usepackage{caption}
\usepackage{graphicx} 
\usepackage{pifont}
\usepackage{multirow}
\usepackage[table]{xcolor}

\definecolor{shenlong}{RGB}{150, 0, 255}

\newcommand{\before}{{\text{pre}}}%
\newcommand{\after}{{\text{post}}}%

\newcommand{\cmark}{\ding{51}}%
\newcommand{\xmark}{\ding{55}}%

\usepackage[pagebackref,breaklinks,colorlinks,citecolor=eccvblue]{hyperref}

\usepackage{orcidlink}

\begin{document}
\title{SceneDiff: A Benchmark and Method for Multiview Object Change Detection} 

\titlerunning{SceneDiff}
\authorrunning{Wu et al.}

\author{%
  Yuqun Wu\inst{1} \qquad
  Chih-hao Lin\inst{1} \qquad
  Henry Che\inst{1} \qquad
  Aditi Tiwari\inst{1} \\
  Chuhang Zou\inst{2}  \qquad
  Shenlong Wang\inst{1} \qquad
  Derek Hoiem\inst{1} \\
}
\institute{%
  $^{1}$University of Illinois at Urbana-Champaign \quad $^{2}$Meta
}



\maketitle

\begin{center}
\begin{minipage}{0.96\linewidth}
\centering
\vspace{-0.2em}
\includegraphics[width=\linewidth]{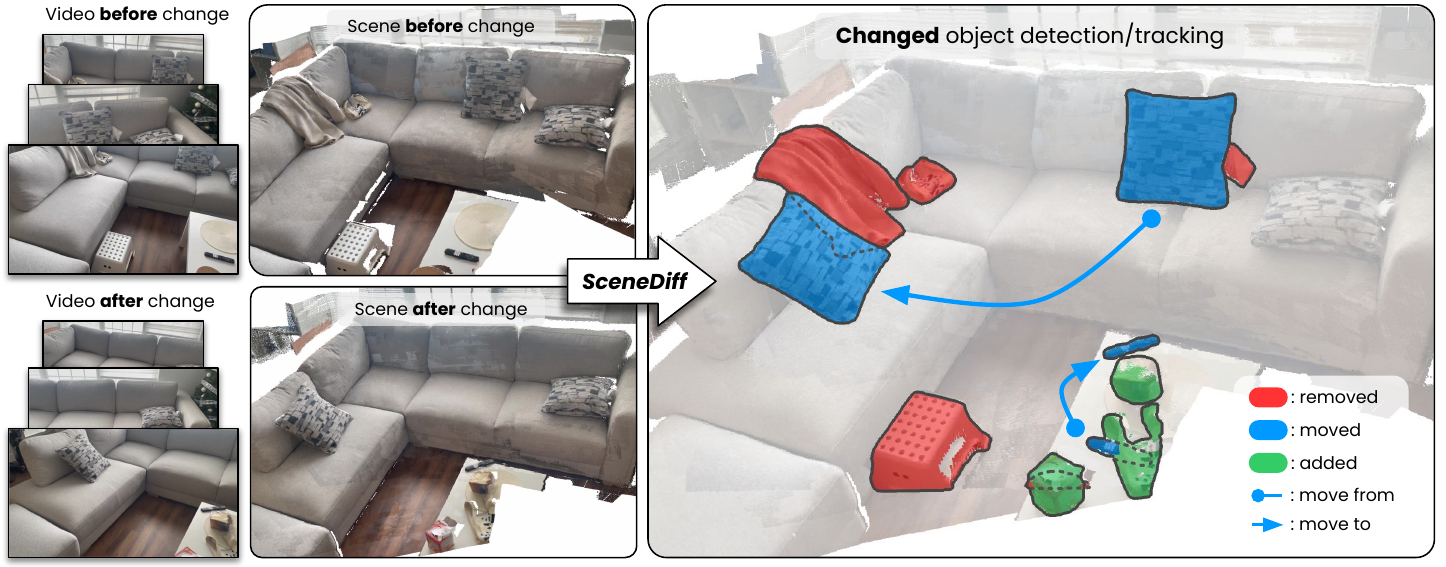}
\vspace{-1.5em}
\captionof{figure}{\textbf{Multiview change detection.} We identify the changed objects (\textcolor[RGB]{255,51,51}{\texttt{Removed}}, \textcolor[RGB]{51,204,102}{\texttt{Added}}, and \textcolor[RGB]{0,153,255}{\texttt{Moved}}) given two videos capturing the same scene at different times. 
The right panel shows a projected 3D visualization of our 2D predictions, with object boundaries manually overlaid. Dashed lines indicate occluded changed objects.}
\label{fig:teaser_figure}
\end{minipage}
\end{center}

\begin{abstract}

We investigate the problem of identifying objects that have been added, removed, or moved between a pair of captures (images or videos) of the same scene at different times. Accurately identifying verifiable changes is extremely challenging -- some objects may appear to be missing because they are occluded or out of frame, while others may appear different due to large viewpoint changes. To study this problem, we introduce the SceneDiff Benchmark, the first multiview change detection dataset for scenes captured along different camera trajectories, comprising 350 diverse video pairs with dense object instance-level annotations. We also introduce the SceneDiff algorithm, a training-free approach that solves for image poses, segments images into objects, and compares them using semantic and geometric features. By building on pretrained models, SceneDiff generalizes across domains without retraining and naturally improves as the underlying models advance. Experiments on multiview and two-view benchmarks demonstrate that our method outperforms existing approaches by large margins (53.0\% and 30.6\% relative AP improvements). Project page: \url{https://yuqunw.github.io/SceneDiff}
\end{abstract}
\section{Introduction}

Object change detection (Fig.~\ref{fig:teaser_figure}), identifying objects that have been added, removed, or moved based on videos or images of the same scene at different times, serves as a fundamental test of spatial understanding. This capability is critical for applications ranging from robotic room tidying (Fig.~\ref{fig:robot_demo}) to construction monitoring, warehouse inventory verification, and post-disaster damage assessment. For example, a superintendent may want to know where new items have been installed; a warehouse operator needs to confirm that pallets are in the correct locations after a shift change; and disaster responders need to catalog what has been displaced or destroyed. However, the task is extremely challenging: significant viewpoint shifts, lighting variations, and occlusions often cause objects to falsely appear changed. To succeed, a method must establish correspondence between the two sets of frames and identify confirmable changes while ignoring apparent changes that are due only to viewpoint or lighting differences.

In this work, we offer SceneDiff Benchmark, the first instance-level multiview change detection dataset designed for scenes captured along different camera trajectories. The dataset contains 350 real-world sequence pairs from 50 diverse scenes across 20 unique scene categories, consisting of 200 manually collected video pairs and 150 egocentric video pairs extracted from the HD-Epic dataset~\cite{Perrett2025HDEPICAH}. Unlike prior benchmarks that assume near-duplicate viewpoints or provide only semantic labels, SceneDiff Benchmark provides dense object instance-level annotations across a large and diverse collection of indoor and outdoor scenes. We provide a SAM2-based~\cite{Ravi2024SAM2S} annotation tool and an evaluation protocol for both per-view and per-scene object change detection assessment.

Analogous to the "diff" command for text files, our approach, SceneDiff, first aligns scene views and then detects differences. Our two key insights are: (1) feed-forward 3D models can use static scene elements to co-register the before and after sequences into a shared 3D space; and (2) once aligned, changed objects produce geometric and appearance inconsistencies that are easier to detect through object-level region features. By aligning scenes and segmenting them into objects before matching, SceneDiff greatly simplifies comparison and improves accuracy relative to methods that operate at the pixel or point level. We leverage pretrained 3D (e.g., $\pi^3$), segmentation (e.g., SAM), and semantic (e.g., DINOv3) foundation models for alignment and comparison, identifying objects with inconsistent appearance or geometry across views. Because SceneDiff is built entirely on pretrained models, it is training-free and straightforward to apply across diverse datasets and domains—handling both image pairs and video sequences—and it naturally improves as the underlying semantic features, 3D representations, or segmentation models advance, as validated in our ablations (Sec.~\ref{sec:5.3}). This approach outperforms prior work by large margins on both our proposed SceneDiff Benchmark (53.0\% relative improvement, averaging obj/im AP in the two portions) and the established two-view RC-3D benchmark~\cite{Sachdeva2023TheCY} (30.6\% relative AP improvement). We demonstrate a robotic application (Sec.~\ref{sec:robot_application}), where our method enables a robot to tidy up a messy table to its original, user-defined setup.

In summary, our key contributions are:
\begin{itemize} 
\item \textbf{SceneDiff Benchmark}, the first multiview dataset for object change detection across different camera trajectories, containing 350 real-world sequence pairs from 50 scenes across 20 scene categories, with dense object instance-level annotations spanning both indoor and outdoor environments. We release the dataset, annotation tool, and evaluation code.
\item \textbf{SceneDiff algorithm}, a training-free framework that co-registers images and segments them into objects before comparison, greatly simplifying matching relative to patch-level approaches. By building entirely on pretrained foundation models, SceneDiff generalizes across domains without retraining and naturally benefits as the underlying models improve. SceneDiff outperforms prior methods on both our proposed benchmark and the established two-view RC-3D benchmark~\cite{Sachdeva2023TheCY}, and we demonstrate its practical utility through a closed-loop robotic tidying application.
\end{itemize}

\section{Related Work}
\textbf{Object change detection} aims to identify objects in the same scene that change over time.
Recent solutions address this problem in terms of two-view~\cite{lin2024robust, park2022dual,Pomerleau2014Longterm3M, Sachdeva2023TheCY, sachdeva2023change, wang2023reduce, chen2021dr, liu2021super, changenet, varghese2025viewdelta}
and multiview~\cite{Wald2019RIO, fu2022robust, looper20233d, sakurada2013detecting, improving0shot} settings. 
Much effort has been devoted to change detection under similar viewpoints, using learning-based~\cite{lin2024robust, ramkumar2021self} or training-free methods~\cite{cho2025zero, Kim_2025_CVPR}.
To address different viewpoints using synthetic training data, CYWS~\cite{sachdeva2023change} learns the differences at the feature level and performs bounding box detection of changes in the scene through a U-Net architecture. Subsequent work~\cite{Sachdeva2023TheCY} warps image features from one image to another via estimated monocular depth.
Other works also train models for dense semantic mask predictions~\cite{park2022dual}.
These methods have demonstrated strong results, though they typically require task-specific training data and can face a sim-to-real gap.
Other explorations assume sensor depth inputs and include better feature representations such as neural descriptor fields~\cite{fu2022robust}, or intermediate representations such as point cloud~\cite{Wald2019RIO,langer2020robust,zhu2023living,bore2019moveable}, 3D semantic scene graph~\cite{looper20233d}, and TSDF~\cite{fehr2017tsdf, schmid2022panoptic, Schmid-RSS24-Khronos,qian2022pocd}. 
Recent work~\cite{huang2023c, jiang2025gaussian, Lu20243DGSCD3G, Galappaththige2024MultiViewPC} uses NeRF~\cite{mildenhall2020nerf} or 3D Gaussian splatting~\cite{Kerbl20233DGS,yugay2025evolvescene} to identify discrepancies between rendered and captured images. 
These render-and-compare approaches effectively leverage 3D consistency, though they perform best when dense input views are available and camera trajectories provide good coverage of the scene.
In contrast, our training-free approach leverages geometry~\cite{wang2025pi3}, semantic~\cite{simeoni2025dinov3}, and segmentation~\cite{Kirillov2023SegmentA} foundation models to robustly detect changes. Furthermore, our proposed benchmark provides a rigorous framework for evaluating these models on the change detection task.

\begin{table*}[t]
\centering
\caption{
    \textbf{Comparison with existing change detection datasets.} 
    The SceneDiff Benchmark is the first dataset with instance-level annotations for video sequences and contains the largest collection of different-viewpoint sequence pairs.
    \textit{Consistent ID} indicates consistent cross-view instance annotation;
    \textit{Viewpoints} indicates whether the before/after viewpoints are similar or different;
    \textit{Out./In.} refer to outdoor/indoor;
    \textit{Data} denotes the raw video or image count; 
    \textit{\# Pairs} denotes the number of frames with ground-truth annotations.
    RC-3D contains one changed object per pair.
}
\vspace{-1em}
\resizebox{0.88\textwidth}{!}{
\begin{tabular}{l|@{\hspace{6pt}}c@{\hspace{6pt}}c@{\hspace{6pt}}c@{\hspace{6pt}}c@{\hspace{6pt}}c@{\hspace{6pt}}r@{\hspace{6pt}}r}
\toprule
Dataset & Consistent ID & Real & Viewpoints & Annotation & Scene & Data & \# Pairs \\
\midrule
VL-CMU-CD~\cite{alcantarilla2018street} & \xmark & \cmark & Similar & Semantic & Out. & 152 Videos & 4.2K \\
PSCD~\cite{sakurada2020weakly} & \xmark & \cmark & Similar & Instance & Out. & 770 Images & 770 \\
ChangeSim~\cite{park2021changesim} & \xmark & \xmark & Similar & Semantic & In. & 80 Videos & 130K \\
3DGS-CD~\cite{Lu20243DGSCD3G} & \xmark & \cmark & Different & Semantic & In. & 5 Videos & 20 \\
PASLCD~\cite{Galappaththige2024MultiViewPC} & \xmark & \cmark & Different & Semantic & Both & 20 Videos & 500 \\
RC-3D~\cite{Sachdeva2023TheCY} & \xmark & \cmark & Different & Instance & In. & 100 Images & 100 \\
\midrule
\textbf{SceneDiff (Ours)} & \cmark & \cmark & \textbf{Different} & \textbf{Instance} & Both & \textbf{350 Videos} & 6K \\
\bottomrule
\end{tabular}
}
\vspace{-1.2em}
\label{tab:dataset_comp}
\end{table*}
\noindent%
\textbf{Change detection benchmarks.} 
Existing change detection benchmarks~\cite{alcantarilla2018street, sakurada2020weakly, park2021changesim, Sachdeva2023TheCY, Lu20243DGSCD3G, Galappaththige2024MultiViewPC} have driven significant progress in the field, primarily focusing on per-view predictions.
Datasets in real-world outdoor~\cite{alcantarilla2018street, sakurada2020weakly} and synthetic indoor environments~\cite{park2021changesim} established pixel-level metrics under the assumption of similar viewpoints.
RC-3D~\cite{Sachdeva2023TheCY} introduces viewpoint variation and provides 100 image pairs, each containing a single changed object.
3DGS-CD~\cite{Lu20243DGSCD3G} and PASLCD~\cite{Galappaththige2024MultiViewPC} provide high-quality semantic change maps for 5 and 20 sequence pairs, which are well-suited for testing change detection under test viewpoints interpolated from densely sampled training views.
To evaluate a method's ability to consistently identify the same physical changed objects across frames, we offer the first dataset for multiview \textit{object-level} change detection, including \textit{350 video pairs} in diverse real-world scenes (Tab.~\ref{tab:dataset_comp}).

\noindent\textbf{Geometry reconstruction} plays a critical role in scene change detection, as accurate geometry supports explicit geometric and appearance cues for identifying changes.
Traditional methods~\cite{yao2018mvsnet,yao2019recurrent,lee2021patchmatch, schoenberger2016mvs,kuhn2020deepc,ma2021eppmvsnet} solve the problem via a two-stage process. 
More recently, DUSt3R~\cite{Wang2023DUSt3RG3} introduces a unified pipeline that directly predicts geometry given two views and shows strong performance. 
Much effort has been invested to further enhance these pipelines for multiview inputs~\cite{Yang2025Fast3RT3, Tang2024MVDUSt3RSS, Wang2025VGGTVG, wang2025pi3, keetha2025mapanything} and dynamic sequences~\cite{zhang2024monst3r, Wang2025Continuous3P, Wang20243DRW}.  
Unlike these works that focus on dynamic scene reconstruction, we aim to solve the scene change detection problem in static scenes captured at different times.
Multiview change detection also serves as a downstream task to evaluate 3D inference models, and we use $\pi^3$~\cite{wang2025pi3} for geometry reconstruction in our method.

\vspace{-0.5em}
\section{SceneDiff Benchmark}
\vspace{-0.4em}


The SceneDiff benchmark evaluates consistent object-level change detection at both per-image and per-scene levels under varying viewpoints (Tab.~\ref{tab:dataset_comp}). 
In this section, we detail the evaluation metrics, dataset statistics, and annotation tool.
\subsection{Metrics}
We offer three metrics for evaluation in SceneDiff, going from per-pixel and view-centric to per-object and scene-centric: (1) \textbf{px/im IoU}: which pixels changed in each image; (2) \textbf{obj/im AP}: which objects changed in each image; (3) \textbf{obj/sc AP}: which objects changed in each scene. 
\textit{px/im IoU} and \textit{obj/im AP} evaluate per-view segmentation quality and detection coverage, following existing benchmarks~\cite{park2021changesim, alcantarilla2018street, sakurada2020weakly, sachdeva2023change}. 
\textit{obj/sc AP} additionally evaluates scene-level object identification consistency across views, requiring the system to determine whether detections across multiple views correspond to the same physical object or to distinct objects. 
For example, if an added chair is detected in two images within the same sequence, does that count as one changed object or two; and if a chair is moved across sequences, is that a moved object or separate removed and added objects?
The superintendent, warehouse operator, or disaster responder is concerned more with objects in the scene than pixels in images, so we consider the third metric our most important, while the other two allow for broader comparison.
\paragraph{Implementation Details.}
For \textit{obj/im AP}, detections in each view are sorted by confidence and matched greedily to ground-truth masks using mask IoU with a threshold of 0.5. Each ground-truth object can only be matched once, and unmatched predictions are counted as false positives.
For \textit{obj/sc AP}, detections across views corresponding to the same object hypothesis are aggregated into a single scene-level prediction, with confidence computed as the mean of per-view prediction confidences. 
A scene-level prediction is matched to a ground-truth object if the IoU aggregated across frames in both sequences exceeds 0.5. More results under different criteria are provided in the supplement.  

\begin{figure*}[t]
\centering
        \includegraphics[width=0.98\linewidth]{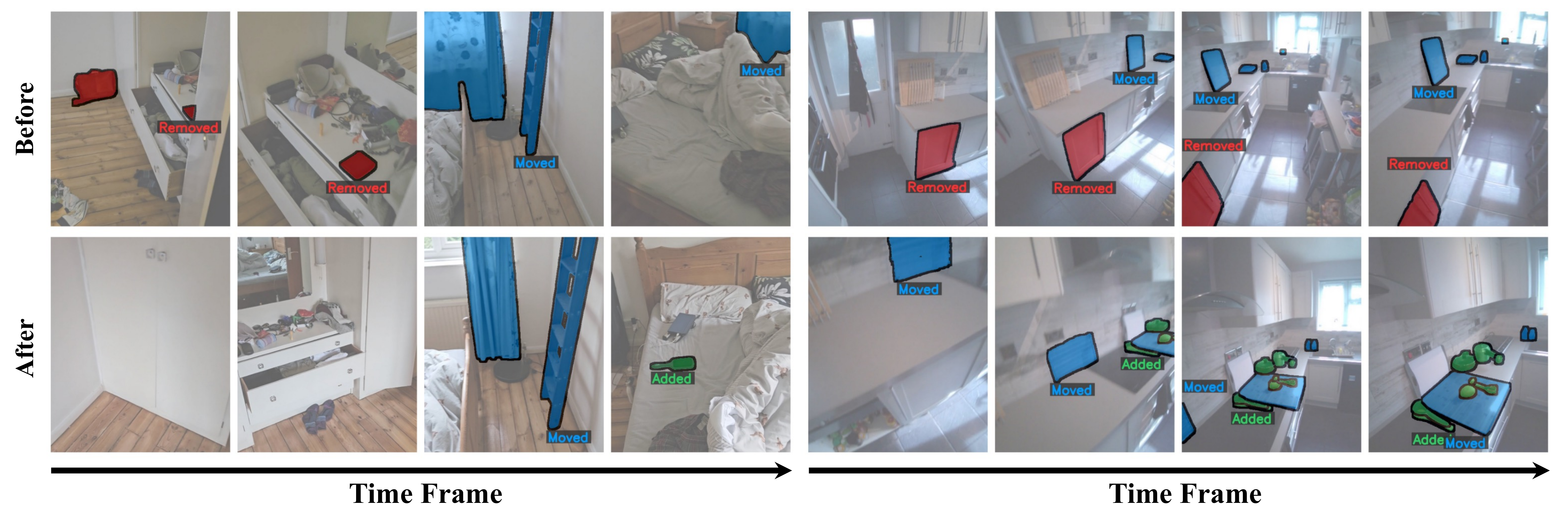}
\vspace{-1.2em}
\caption{
    \textbf{Sequence Examples.} We visualize video pairs before and after changes. 
    Changed objects are color-masked by change type: \textcolor[RGB]{255,51,51}{\texttt{Removed}}, \textcolor[RGB]{51,204,102}{\texttt{Added}}, and \textcolor[RGB]{0,153,255}{\texttt{Moved}}. The background is masked white.
    The first pair is from \textit{SD-V}, and the second is from \textit{SD-K}. 
    }
\label{fig:multi_view}
\vspace{-.8em}
\end{figure*}

\begin{figure*}[t]
\centering
        \includegraphics[width=0.99\linewidth]{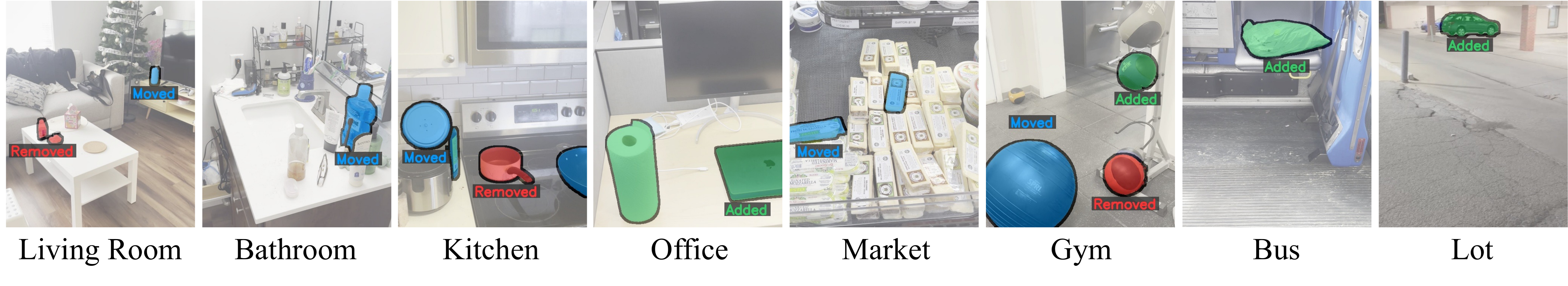}
\vspace{-.8em}
\caption{
    \textbf{Diverse Scene Examples.} Annotated frames from the SceneDiff Benchmark. 
}
\label{fig:scene_example}
\vspace{-2em}
\end{figure*}

\subsection{Dataset}
\begin{figure*}[t]
\centering
        \includegraphics[width=0.99\linewidth]{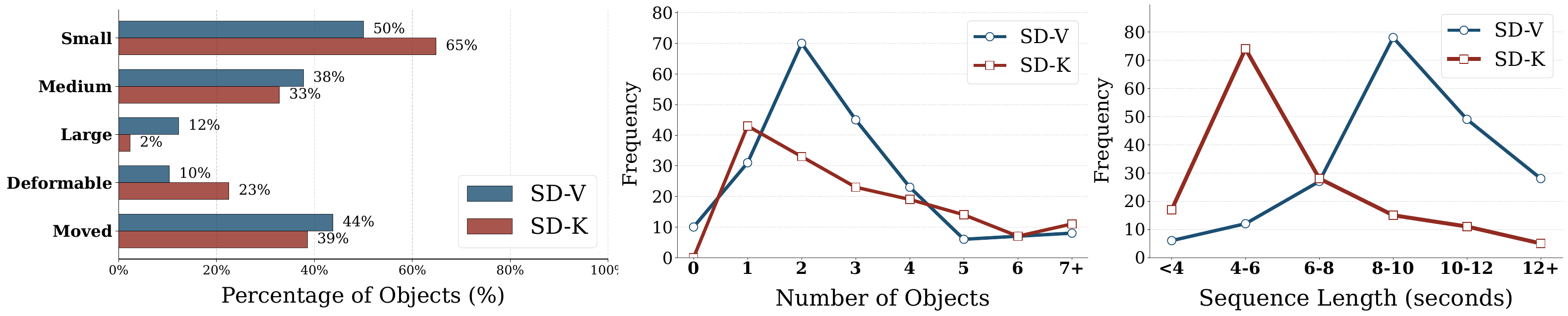}
\vspace{-1em}
\caption{
    \textbf{Dataset Statistics.} Distribution of object properties, changed object counts, and sequence lengths in the SceneDiff Benchmark. Object size categorization is based on the average pixel size across all frames. \textit{SD-V} contains larger objects and longer sequences, while \textit{SD-K} contains more deformable objects.
}
\label{fig:dataset_stat}
\vspace{-1.5em}
\end{figure*}

SceneDiff (Fig.~\ref{fig:multi_view}) contains $350$ video sequence pairs and $1009$ annotated objects across the \textit{Varied} and \textit{Kitchen} subsets. 
The \textit{Varied} subset~(\textit{SD-V}) contains 200 sequence pairs collected by 7 people in a wide variety of daily indoor and outdoor scenes (Fig.~\ref{fig:scene_example}), and the \textit{Kitchen} subset~(\textit{SD-K}) contains 150 sequence pairs from the HD-Epic dataset~\cite{Perrett2025HDEPICAH} with changes that naturally occur during cooking activities. 
For each video pair, we record all changed object attributes, including object names, change types (\texttt{Added}, \texttt{Removed}, or \texttt{Moved}), and deformability, and annotate their full segmentation masks in all visible frames.
In total, the sequence collection, tool development, and annotation required approximately 160, 30, 60 hours respectively.
To avoid ambiguity, we avoid capturing objects that are moving within a single sequence, so that all changes are across sequences.
An overview of the dataset characteristics is presented in Fig.~\ref{fig:dataset_stat}, including the distribution of changed objects per video, duration of video sequences, and object attributes (size, deformability, and category).
Our validation set, intended for training or hyperparameter tuning, is 50 video pairs per subset, and the test set is the remaining 250 pairs, with all pairs from the same physical scene placed in the same split.

\vspace{-0.5em}
\subsection{Annotation Tool}
\vspace{-0.4em}

Annotating dense object masks across video pairs is time-consuming, even with modern tools such as SAM2~\cite{Ravi2024SAM2S}. 
To streamline this process, we develop an annotation interface based on SAM2 in which users upload video pairs, specify object attributes (deformability, change type, multiplicity), and provide sparse point prompts on selected frames via clicking. 
The system records these prompts, propagates masks throughout both videos offline, and provides a review interface where users can visualize the annotated videos, refine annotations if needed, and submit verified pairs to the dataset. 
This reduces annotation time from approximately 30 minutes to 10 minutes per video pair. 
The annotation video and other details are in the supplement.
\begin{figure*}[t]
\centering
        \includegraphics[width=0.98\textwidth]{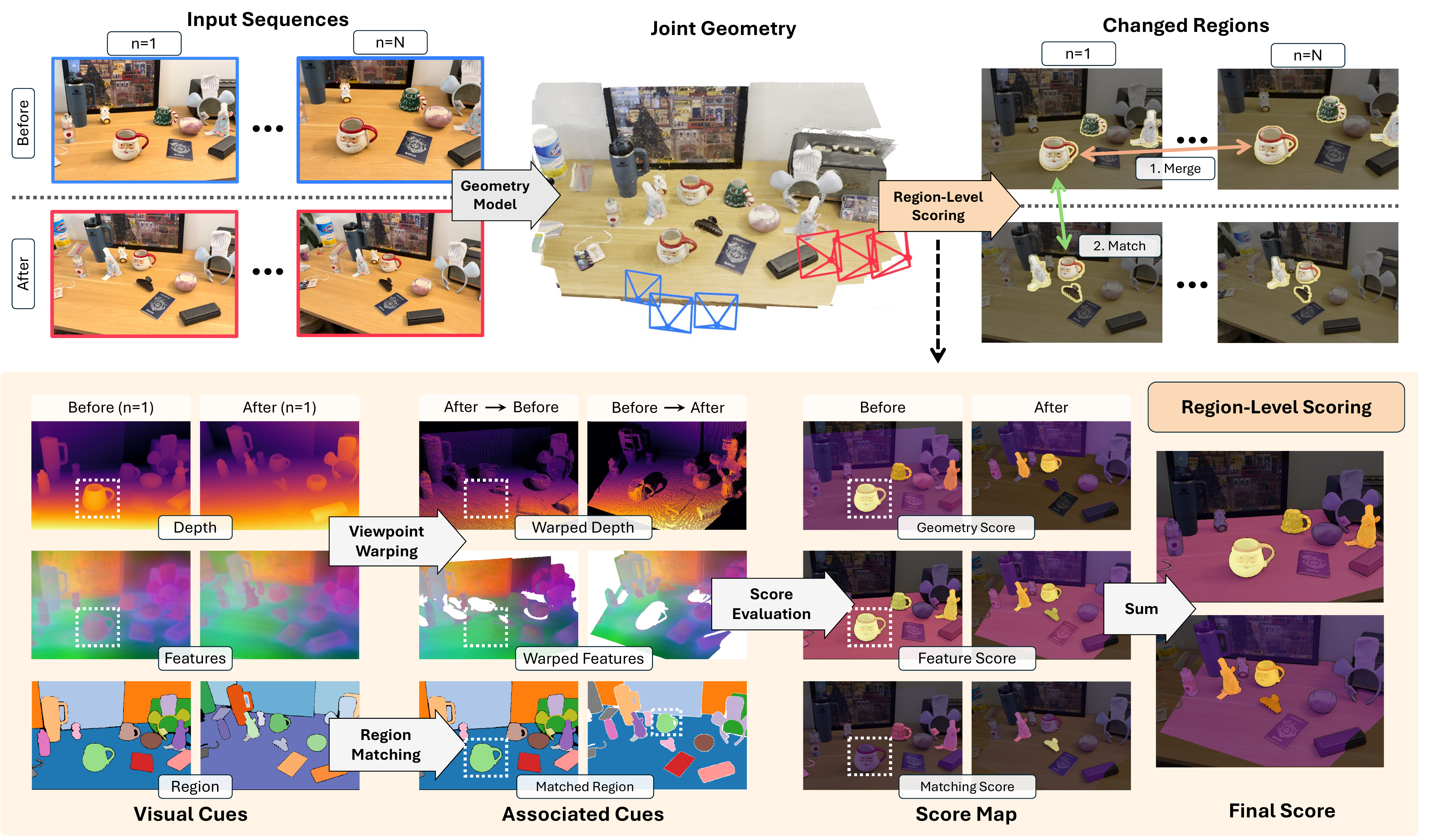}
\vspace{-0.6em}
\caption{ \textbf{SceneDiff Method.} \textit{Top (Overall Approach):} Given before and after sequences, SceneDiff jointly regresses geometry, selects paired views with high co-visibility, and computes region-level change scores. Changed regions are then merged within each sequence into object-level changes and matched across sequences to classify the change type. 
\textit{Bottom (Region-Level Change Scoring):} For each paired view, we compute three scores: a geometry score via depth reprojection, an appearance score via feature reprojection, and a region matching score by comparing mean-pooled features across images. These are combined and pooled to produce unified region-level change scores. }
\vspace{-1.2em}
\label{fig:architecture}
\end{figure*}
\vspace{-0.5em}
\section{SceneDiff Method}
\vspace{-0.5em}
SceneDiff aims to detect the changed objects given two image sequences taken before and after the scene changes~(Fig.~\ref{fig:architecture}). 

\noindent \textbf{Task Definition:} 
Given two videos, $\mathcal{I}_\before = \{\mathbf{I}_\before^n\}_{n=1}^{N_\before}$ and $\mathcal{I}_\after = \{\mathbf{I}_\after^n\}_{n=1}^{N_\after}$, captured before and after the change, our goal is to identify verifiable object-level changes across all views, while ignoring apparent differences caused solely by viewpoint variations, e.g., occlusion, partial visibility, and different illumination.
The output is a set of changed objects with corresponding 2D masks in all visible views and their change type: \textit{Added} (present only in $\mathcal{I}_\after$), \textit{Removed} (present only in $\mathcal{I}_\before$), or \textit{Moved} (present in both but at different positions).
When $N_\before = N_\after = 1$, the task reduces to two-view change detection.

\noindent\textbf{Overall Approach:} 
Scene changes produce geometric and appearance discrepancies in 3D space. 
However, viewpoint variations can also create differences, making these cues noisy.
To address this, SceneDiff first selects view pairs with high co-visibility from the before and after videos to ensure comparable perspectives.
It then computes region-level change scores from geometry and appearance cues for each view pair, detects changed regions, associates these regions across time to merge them into consistent object-level changes, and finally produces object-level change segmentation (2D and 3D) and classification.

\vspace{-0.5em}
\subsection{Geometry Regression and Frame Pairing}
\label{view_pairing}
We process the video pair---$\mathcal{I}_\before$ and $\mathcal{I}_\after$ jointly through $\pi^3$~\cite{wang2025pi3} to estimate depth, pose, and intrinsics, yielding \{$\mathbf{D}_\before^n, \mathbf{T}_\before^n, \mathbf{K}_\before^n\}$ and $\{\mathbf{D}_\after^n, \mathbf{T}_\after^n, \mathbf{K}_\after^n\}$, where $\{\cdot\}$ denotes a set over frame index $n$.
To ensure a consistent scale across scenes, we normalize $\{\mathbf{D}_\before^n, \mathbf{D}_\after^n\}$ and $\{\mathbf{T}^n_\before, \mathbf{T}^n_\after\}$ so that the reconstructed point clouds fit within a unit cube $[-1, 1]^3$.


We then select frames across ``before'' and ``after'' sequences for change detection.
Not all frame pairs are suitable for matching due to camera motion and limited overlap.
Therefore, for each frame $n$ in the before sequence, we select one or more frames $n'$ in the after sequence, considering $(\mathbf{I}_\before^n,\mathbf{I}_\after^{n'})$ a good pair if (1) their \emph{bidirectional co-visibility}—the average fraction of mutually visible pixels under before\,$\leftrightarrow$\,after reprojection—exceeds $50\%$, or (2) $\mathbf{I}_\after^{n^\prime}$ has the highest co-visibility among all frames in $\mathcal{I}_\after$. 
The selected pairs are then used to compute change scores; for simplicity, we denote a paired view as $(\mathbf{I}_\before,\mathbf{I}_\after)$.

\subsection{Region-Level Change Scoring}
Reprojected depth and appearance differences may reveal scene changes, but two challenges exist: 1) differences may arise from viewpoint variation, e.g., occlusion, rather than actual scene changes; 2) pixel-level discrepancies contain noise and cannot capture object-level changes.
To address these issues, we use reprojected depth to distinguish true scene changes from viewpoint-dependent occlusions, leverage DINOv3~\cite{simeoni2025dinov3} features for robust appearance comparison, and aggregate pixel-level cues into region-level scores using masks predicted by SAM~\cite{Kirillov2023SegmentA}.

\vspace{1mm}
\noindent \textbf{Geometry Comparison:}
When reprojecting pixels from $\mathbf{I}_\before$ to $\mathbf{I}_\after$, objects that \textbf{exist only in} $\mathbf{I}_\before$ yield a reliable cue: the observed depth value in $\mathbf{I}_\after$ (background) is larger than that of reprojected depth (object), producing positive differences. In contrast, negative differences are ambiguous, as they can arise from occlusion or from objects that \textbf{exist only in} $\mathbf{I}_\after$. 
Accordingly, we adopt an \emph{asymmetric} rule: we only use pixels in $\mathbf{I}_\before$ with non-negative depth differences when reprojected to $\mathbf{I}_\after$ to detect changes visible in $\mathbf{I}_\before$; by reversing the view order, we detect changes visible in $\mathbf{I}_\after$.

Specifically, for a pixel $p$ with coordinates $(x_p,y_p)$ in $\mathbf{I}_\before$, we first transform it into the camera space of $\mathbf{I}_\after$ as
$\mathbf{p}^{3d}_\after = \mathbf{T}_\after^{-1}\mathbf{T}_\before\,\mathbf{D}_\before(p)\mathbf{K}_\before^{-1}[x_p, y_p, 1]^\top$,
and compute the reprojected pixel 
$p' = \pi(\mathbf{K}_\after \mathbf{p}^{3d}_\after)$ 
and the corresponding reprojected depth
$\mathbf{D}_\after^{\ast}(p') = [\mathbf{p}^{3d}_\after]_z$,
where $\pi(\cdot)$ denotes projection to image space and $[\cdot]_z$ denotes the $Z$-component.
We then define the depth-difference score $\mathbf{E}_{\text{geom}}$:
\begin{equation}
\mathbf{E}_{\text{geom}}(p) = \mathbf{D}_\after(p') - \mathbf{D}_\after^\ast(p').
\end{equation}
As described, positive values indicate objects only appear in $\mathbf{I}_\before$; negative values reflect occlusion or objects only appear in $\mathbf{I}_\after$. 
To enforce the asymmetric rule, we construct a directional visibility mask that accounts for non-negative differences and fields of view: $\mathbf{M}_\before = \big(\mathbf{E}_{\text{geom}} \ge \tau_{\text{occ}}\big) \wedge \mathbf{V}_{\before\to\after}$
, where $\tau_{\text{occ}}=-0.02$ and $\mathbf{V}_{\before\to\after}$ is the visibility mask from the $\before\!\to\!\after$ reprojection. 
We swap the view order to obtain $\mathbf{M}_\after$ for detecting changes visible in $\mathbf{I}_\after$.

\vspace{1mm}
\noindent 
\textbf{Appearance Comparison:}
To detect appearance changes, we extract DINOv3~\cite{simeoni2025dinov3} features ($\mathbf{F}_\before$, $\mathbf{F}_\after$) and measure feature dissimilarity via reprojection. 
We then acquire the reprojected appearance features $\mathbf{F}_\before^{\ast}$ from $\mathbf{F}_\after$, apply the directional visibility mask $\mathbf{M}_\before$ to exclude the invisible areas, and obtain the reprojected feature score $\mathbf{E}_{\text{feat}}$. Specifically, given pixel $p$ in $\mathbf{I_\before}$ and the corresponding pixel $p'$ in $\mathbf{I_\after}$:
\begin{equation}
\mathbf{E}_{\text{feat}}(p)=\mathbf{M}_\before(p)\Bigl(1-\cos(\mathbf{F}_\before(p),\mathbf{F}_\before^{\ast}(p))\Bigr),
\end{equation}
where $\mathbf{F}_\before^{\ast}(p) = \mathbf{F}_\after(p')$ is the reprojected feature of $p$.

\vspace{1mm}
\noindent \textbf{Region-Level Matching:}
In addition to $\mathbf{E}_{\text{feat}}$, which compares reprojected features pixelwise and then pools over regions, we propose region-level matching, that first pools features and then compares regions at any position.
We extract regions ($\mathcal{R}_\before$, $\mathcal{R}_\after$) from SAM~\cite{Kirillov2023SegmentA}, where each region $r \in \mathcal{R}$ corresponds to a set of pixels from a predicted mask. 
Reprojection-based cues ($\mathbf{E}_{\text{geom}}$ and $\mathbf{E}_{\text{feat}}$) heavily rely on accurate geometry. To enhance robustness, we additionally measure whether regions have corresponding objects in the other view based on appearance alone. 
We aggregate features $\mathbf{F}_\before$ within each region defined by masks $\mathcal{R}_\before$ and compute the mean-pooled feature vector $\mathbf{F}_\before^{r} = \frac{1}{|r|}\sum_{p \in r} \mathbf{F}_\before(p)$ for each region $r$ in $\mathbf{I}_\before$.
We then use cosine similarity to select the best-matching region $\sigma(r)$ in $\mathbf{I_{post}}$ and compute the region matching score $E_{\text{region}}$ over $r$ as: 
\begin{equation} 
\begin{aligned} 
E_{\text{region}}(r) &= 1 - \cos(\mathbf{F}_\before^{r}, \mathbf{F}_\after^{\sigma(r)}), 
\end{aligned} 
\end{equation} 
where $\sigma(r) = \arg\max_{s \in \mathcal{R}_\after} \cos(\mathbf{F}_\before^{r}, \mathbf{F}_\after^{s})$.
To account for occlusion and visibility, we exclude regions with more than 50\% pixels masked by $\mathbf{M}_\before$ from this matching process.

\noindent \textbf{Score Aggregation:}
All cost maps are mean-pooled with the region masks, and combined with a weighted sum to obtain the unified score map $\Delta_\before$ over each $r \in \mathcal{R}_\before$:
\begin{equation}  
\Delta_\before(r) = \frac{\lambda^{geom}}{|r|}{\sum_{p \in r} \mathbf{E}_{\text{geom}}(p)} + \frac{\lambda^{feat}}{|r|}{\sum_{p \in r} \mathbf{E}_{\text{feat}}(p)} + \lambda^{region} E_{\text{region}}(r) 
\end{equation}
where $|r|$ is the number of pixels in region $r$
and $\lambda^{geom}=1.0$, $\lambda^{feat}=0.5$, and $\lambda^{region}=0.1$ are score weights. 
Similarly, we compute $\Delta_\after$ by swapping $\mathbf{I}_\before$ and $\mathbf{I}_\after$ for all regions in $\mathcal{R}_\after$. 

\subsection{Instance Association and Change Classification}
We use the change scores to retrieve the changed regions in each frame. However, the detected per-frame regions across different frames can correspond to the same object. 
Practically, we often want to know how many object instances changed, not how many regions changed. This requires identifying which regions across frames correspond to the same object instance.

\vspace{1mm}
\noindent \textbf{Frame-Level Change Detection:}
For each view $\mathbf{I}_\before^n$, we compute the averaged unified score maps $\bar{\Delta}_\before^n$ across all its matched frame pairs.
For 3D consistency, we unproject all averaged score maps $\bar{\Delta}_\before^n, \forall n$ into 3D, voxelize the resulting point clouds, average scores within each voxel, and mean-pool again for region-level score maps. 
Given updated \{$\bar{\Delta}_\before^n$\} and \{$\bar{\Delta}_\after^n$\}, we apply a threshold $\tau_{\Delta}$, and retrieve a set of changed regions $\mathcal{R}_\text{before}^{\triangle} = \bigcup_{n=1}^{N_\before} \{r \in \mathcal{R}_\text{before}^{n} \mid \bar{\Delta}_\before^n(r) > \tau_{\Delta}\}$, with each region $r$ associated with a change score $\bar{\Delta}_\before^n(r)$. 
To choose the matching threshold $\tau_{\Delta}$, we use the maximum entropy thresholding algorithm~\cite{kapur1985new}, which works as well as picking the single best threshold on the validation set, but applies more easily to other datasets.

\vspace{1mm}
\noindent \textbf{Video-Level Region Association:}
To obtain instance-level changes, we merge regions across frames using an iterative procedure inspired by ConceptGraph~\cite{Gu2023ConceptGraphsO3}. A region is considered another view of an existing object if their features and 3D points are similar.
Specifically, we initialize our object set $\mathcal{O}_\before$ with $\mathcal{R}_\before^{0, \triangle}$; then for each frame $n$, we iteratively measure the similarity score between each region $r \in \mathcal{R}_\before^{n, \triangle}$ and the changed objects $o \in \mathcal{O}_\before$:

\begin{equation}
\label{eq:region_merge}
\begin{aligned}
    S(r) &= \max_{o \in \mathcal{O}_\before} \left( S_{\text{feat}}(o, r) + S_{\text{geo}}(o, r) \right),
\end{aligned}
\end{equation}
where $S_{\text{feat}}(o, r) = \cos(\mathbf{F}_\before^r, \mathbf{F}_\before^o)$ measures appearance feature similarity between the region and
object; and $S_{\text{geo}}(o, r) = \sum_{\mathbf{x} \in \mathcal{P}_\before^r} \mathbf{1}\left(\min_{\mathbf{y} \in \mathcal{P}_\before^o} \| \mathbf{x} - \mathbf{y} \|_2^2 < \sigma_{\text{geo}} \right) /$\  $|\mathcal{P}_\before^r|$ measures region-to-object geometry similarity as the fraction of region’s points close enough to the object’s point cloud with $\sigma_{\text{geo}} = 0.02$.
If $S(r)$ is higher than the merging threshold~($\sigma_{\text{merge}} = 1.4$), we merge region $r$ into the most similar object $o$, updating the object’s feature via a running average and its point cloud via concatenation: 
\begin{equation}
\mathbf{F}_\before^o \leftarrow w^o\,\mathbf{F}_\before^r + (1 - w^o)\,\mathbf{F}_\before^o, \mathcal{P}_\before^o \leftarrow \mathcal{P}_\before^o \cup \mathcal{P}_\before^r
\end{equation}
where $w^o = 1/(N^o+1)$ and $N^o$ is the number of regions merged into $o$. Otherwise, we instantiate a new object 
$o^\prime$ with $\mathbf{F}_\before^{o^\prime} \leftarrow \mathbf{F}_\before^r$ and $\mathcal{P}_\before^{o^\prime} \leftarrow \mathcal{P}_\before^r$. This process provides a set of changed objects $\mathcal{O}_\before$. We compute the changed objects $\mathcal{O}_\after$ in the second video following the same process. 

\vspace{1mm}
\noindent \textbf{Object Status:}
We perform one-to-one greedy matching between objects in $\mathcal{I}_\before$ and $\mathcal{I}_\after$. 
Pairs with feature cosine similarity greater than $\tau_{\text{sim}} = 0.7$ are sorted in descending order and matched sequentially.
Matched objects are labeled as \textit{Moved} and unmatched objects are labeled as \textit{Added} (if in $\mathcal{I}_\after$) or \textit{Removed} (if in $\mathcal{I}_\before$).



\vspace{1mm}
\noindent \textbf{Parameters}: Feature weight and threshold parameters are set to the indicated values based on the SceneDiff validation set, with the same parameters used across all benchmarks. The supplemental includes sensitivity analysis of thresholds. 

\subsection{Two-Image Input Case}

There may be only one before image and one after image, instead of two videos. In that simpler case, we skip 
3D consistency aggregation and video-level region association,
treating each detected region as an independent object and applying the change type classification directly at the region level with the same parameters.

\vspace{-0.5em}
\section{Experiments}
\vspace{-0.5em}
We show results on our new SceneDiff benchmark (Sec.~\ref{sec:5.1}) and on the two-view change detection dataset (Sec.~\ref{sec:5.2}). In Sec.~\ref{sec:5.3}, we ablate key design choices and the impact of using different features and geometry estimation models. 
We also demonstrate a robotic application in Sec.~\ref{sec:robot_application}.
More results, e.g., failure cases, robustness of geometry models under change, predictions with dynamic contents, time analysis, robustness with different hyperparameters, are provided in Supp.

\begin{table*}[t]
\centering
\caption{
\textbf{Multiview Change Detection on SceneDiff test set.}
We report metrics that correspond to each method's output type: for pixel masks, px/im IoU; for bounding boxes, obj/im AP; for object-level multiview tracks, obj/sc AP.  Bounding boxes are converted to masks using SAM for mask-based obj/im AP.  Object-level tracks are converted to per-image instance masks or pixel masks for comparison.
$^\ast$ indicates results after finetuning on our validation set.
}
\vspace{-1.em}

\newcommand{\gcell}{\cellcolor{gray!15}-}

\fontsize{7.5}{8.5}\selectfont
\setlength{\tabcolsep}{2pt} 
\renewcommand{\arraystretch}{1.1} 

\begin{tabular}{c|c|c|ccc|ccc}
\toprule
 & Output Type & \textbf{Pixel} & \multicolumn{3}{c|}{\textbf{Bounding Box}} & \multicolumn{3}{c}{\textbf{Object Region Tracks}} \\
\cmidrule(lr){3-3} \cmidrule(lr){4-6} \cmidrule(lr){7-9}
 & Method & MV3DCD & CYWS-3D & CYWS-2D & CYWS-2D$^\ast$ & VLM & 3DGS-CD & \textbf{SceneDiff} \\
\midrule
\multirow{3}{*}{\rotatebox{90}{\textbf{SD-V}}}
& \textit{px/im IoU}  &  22.1 & \gcell & \gcell & \gcell & 22.9 & 14.7 & \textbf{39.1} \\
& \textit{obj/im AP}  & \gcell & 13.9 & 19.4 & 24.5 & 7.1  & 1.0  & \textbf{43.8} \\
& \textit{obj/sc AP}  & \gcell & \gcell & \gcell & \gcell & 5.3  & 0.5  & \textbf{22.8} \\
\midrule
\multirow{3}{*}{\rotatebox{90}{\textbf{SD-K}}}
& \textit{px/im IoU}  &  9.1  & \gcell & \gcell & \gcell & 12.4 & 4.8  & \textbf{20.8} \\
& \textit{obj/im AP}  & \gcell & 6.6  & 9.0  & 16.8 & 3.6  & 0.1  & \textbf{20.9} \\
& \textit{obj/sc AP}  & \gcell & \gcell & \gcell & \gcell & 2.1  & 0.1  & \textbf{10.6} \\
\bottomrule
\end{tabular}

\vspace{-0.8em}
\label{tab:two_set_tab}
\end{table*}
\begin{figure*}[t]
\centering
\hspace{-2em}
        \includegraphics[width=1.0\textwidth]{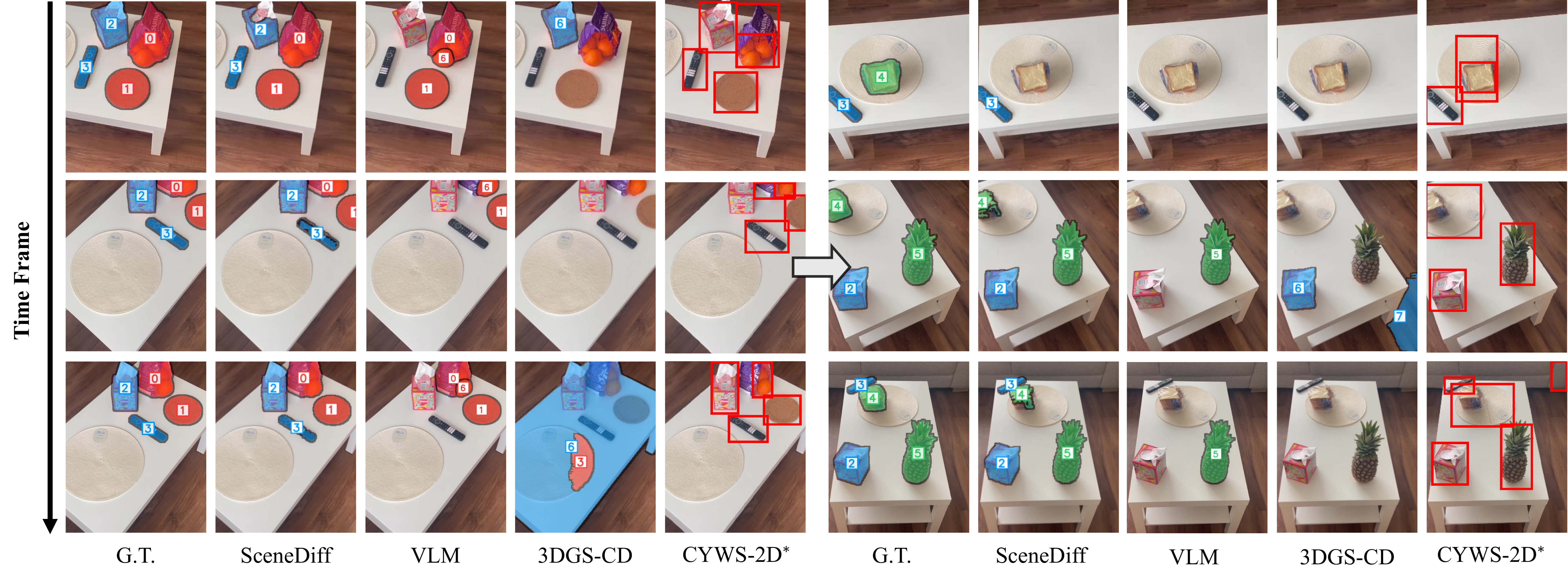}
\vspace{-1.2em}
\caption{
\textbf{Qualitative comparison on the SceneDiff benchmark.}
G.T. objects are visualized with numbers. Object-level predictions show matched GT IDs, or unique IDs if unmatched. Our method correctly predicts all changed objects but misses the bread in a single view. 3DGS-CD produces some correct per-view detections but struggles to associate them into consistent objects. The VLM baseline names accurate but incomplete changed objects (\textit{``Removed: orange, snack bag, coaster; Added: pineapple, sandwich''}). 
Finetuned CYWS-2D successfully detects most changes (showing top-5 predictions after NMS). Color map: \textcolor[RGB]{255,51,51}{\texttt{Removed}}, \textcolor[RGB]{51,204,102}{\texttt{Added}}, and \textcolor[RGB]{0,153,255}{\texttt{Moved}}.
}
\label{fig:qualitative}
\vspace{-1.9em}
\end{figure*}
\subsection{Two-Sequence Change Detection}
\label{sec:5.1}
\begin{figure*}[t]
\centering
        \includegraphics[width=0.95\textwidth]{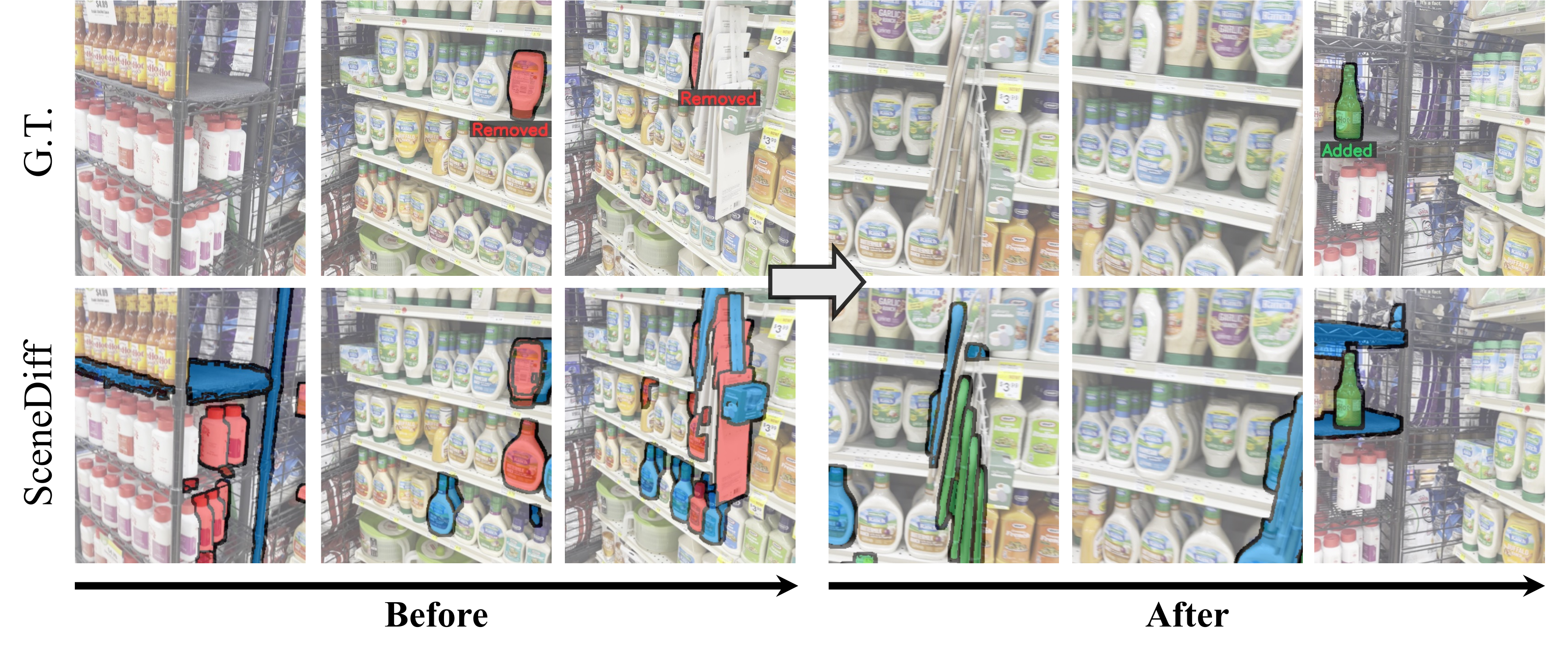}
\vspace{-1.5em}
\caption{
    \textbf{A more challenging sequence pair.} Color map:
    \textcolor[RGB]{255,51,51}{\texttt{Removed}}, \textcolor[RGB]{51,204,102}{\texttt{Added}}, and \textcolor[RGB]{0,153,255}{\texttt{Moved}}. }
\vspace{-.5em}    
\label{fig:qualitative_ours}
\end{figure*}

We present the evaluation results in Tab.~\ref{tab:two_set_tab}, comparing SceneDiff against pair-wise detection methods~\cite{Sachdeva2023TheCY, sachdeva2023change}, 3D Gaussian Splatting (3DGS) based approaches~\cite{Galappaththige2024MultiViewPC, Lu20243DGSCD3G}, and a vision-language model (VLM) baseline.
SceneDiff outperforms all baselines across all metrics, especially in \textit{obj/sc AP}.
However, there is much room for improvement, particularly for the \textit{SD-K} subset, which features complex and cluttered scenes from an egocentric viewpoint.
For pair-wise detection methods CYWS-2D and CYWS-3D~\cite{sachdeva2023change, Sachdeva2023TheCY}, we apply our view-pairing step (Sec.~\ref{view_pairing}) and convert box predictions to masks using SAM~\cite{Kirillov2023SegmentA}. We finetune CYWS-2D on 3125 frame pairs from the validation set and use the official pretrained model for CYWS-3D (training code not released).
3DGS based methods~\cite{Galappaththige2024MultiViewPC, Lu20243DGSCD3G} rely on accurate novel view rendering for change detection. However, high-quality rendering is challenging for current 3DGS methods given large viewpoint changes across trajectories, even when we use our regressed camera parameters as input and sample more densely (3 FPS) for 3DGS optimization. 
We follow 3DGS-CD~\cite{Lu20243DGSCD3G} by using SAM to assign the symmetric change prediction from MV3DCD~\cite{Galappaththige2024MultiViewPC} to the correct sequence.
We also evaluate SceneDiff on PASLCD~\cite{Galappaththige2024MultiViewPC}, outperforming the best baseline by 1.2 mIoU points (Tab.~\ref{tab:paslcd_tab}).
For the VLM baseline, we use Qwen2.5-VL~\cite{Qwen2.5-VL} to predict changed object names and types from both sequences, followed by SAM3~\cite{Carion2025SAM3S} for localization.
For AP computation, we use SAM3’s mask confidence as the detection confidence, since VLMs generate probabilities that are not calibrated for instance-level ranking.
To isolate the VLM's recognition capability from localization errors, we manually match predicted text outputs to ground-truth labels: the VLM correctly names 113/369 changed objects in SD-V (1837 total predictions) and 42/327 in SD-K (792 total predictions), indicating bottlenecks in both recognition and localization.
Fig.~\ref{fig:qualitative} shows the qualitative comparisons, and Fig.~\ref{fig:qualitative_ours} highlights a challenging case, illustrating that this task remains far from solved. 
See Supp. for more details and analysis. 

\begin{table*}[t]
\centering
\providecommand{\gcell}{\cellcolor{gray!15}-}

\begin{minipage}[t]{0.48\linewidth}
    \centering
    \caption{
    \textbf{Change Detection on PASLCD~\cite{Galappaththige2024MultiViewPC}.} We report mIoU for pixel-level symmetric change prediction across indoor and outdoor environments. 
    Despite not being the primary focus, SceneDiff generalizes well and achieves competitive results.
    Baselines are sourced from MV3DCD~\cite{Galappaththige2024MultiViewPC}. See full results in Supp.
    }
    \vspace{-0.8em}
    \renewcommand{\arraystretch}{1.15}

    \resizebox{0.95\linewidth}{!}{ 
    \begin{tabular}{l@{\hspace{6pt}}|@{\hspace{6pt}}c@{\hspace{6pt}}c@{\hspace{6pt}}c}
    \toprule
    \textbf{Method}
    & \textbf{Mean} $\uparrow$ 
    & \textbf{Indoor} $\uparrow$ 
    & \textbf{Outdoor} $\uparrow$ \\
    \midrule
    CSCDNet~\cite{sakurada2020weakly} & 12.5  & 11.7 & 13.4  \\
    CYWS-2D~\cite{sachdeva2023change} & 27.3  & 32.9 & 21.7  \\
    MV3DCD~\cite{Galappaththige2024MultiViewPC} & 46.1  & 44.9 & 47.4  \\
    \textbf{SceneDiff} & \textbf{47.3} & \textbf{46.5} & \textbf{48.0}\\
    \bottomrule
    \end{tabular}
    }
    \label{tab:paslcd_tab}
\end{minipage}\hfill 
\begin{minipage}[t]{0.48\linewidth}
    \centering
    \caption{
    \textbf{Two-View Change Detection on RC-3D~\cite{Sachdeva2023TheCY}.} We report AP50 for box predictions across \textit{Present}, \textit{Absent}, and \textit{Both} views. 
    \textit{Blank entries denote metrics unavailable in the original paper or public codebase.} 
    CYWS-2D/3D baselines are sourced from the original paper. $^\ast$ denotes results incorporate sensor depth.
    }
    \vspace{-0.8em}
    \renewcommand{\arraystretch}{1.15}

    \resizebox{0.95\linewidth}{!}{ 
    \begin{tabular}{l@{\hspace{6pt}}|@{\hspace{6pt}}c@{\hspace{6pt}}c@{\hspace{6pt}}c}
    \toprule
    \textbf{Method}
    & \textbf{Both} $\uparrow$ 
    & \textbf{Present} $\uparrow$ 
    & \textbf{Absent} $\uparrow$ \\
    \midrule
    CYWS-2D~\cite{sachdeva2023change} & 14.0 & \gcell & \gcell \\
    CYWS-3D~\cite{Sachdeva2023TheCY} & 41.0 & \gcell & \gcell \\
    CYWS-3D$^\ast$~\cite{Sachdeva2023TheCY} & 50.0 & \gcell & \gcell \\
    \textbf{SceneDiff} & \textbf{65.3} & \textbf{72.3} & \textbf{59.1} \\
    \bottomrule
    \end{tabular}
    }
    \label{tab:two_view_tab}
\end{minipage}

\vspace{-1.5em}
\end{table*}

\begin{figure*}[h]
\centering
        \includegraphics[width=1.0\textwidth]{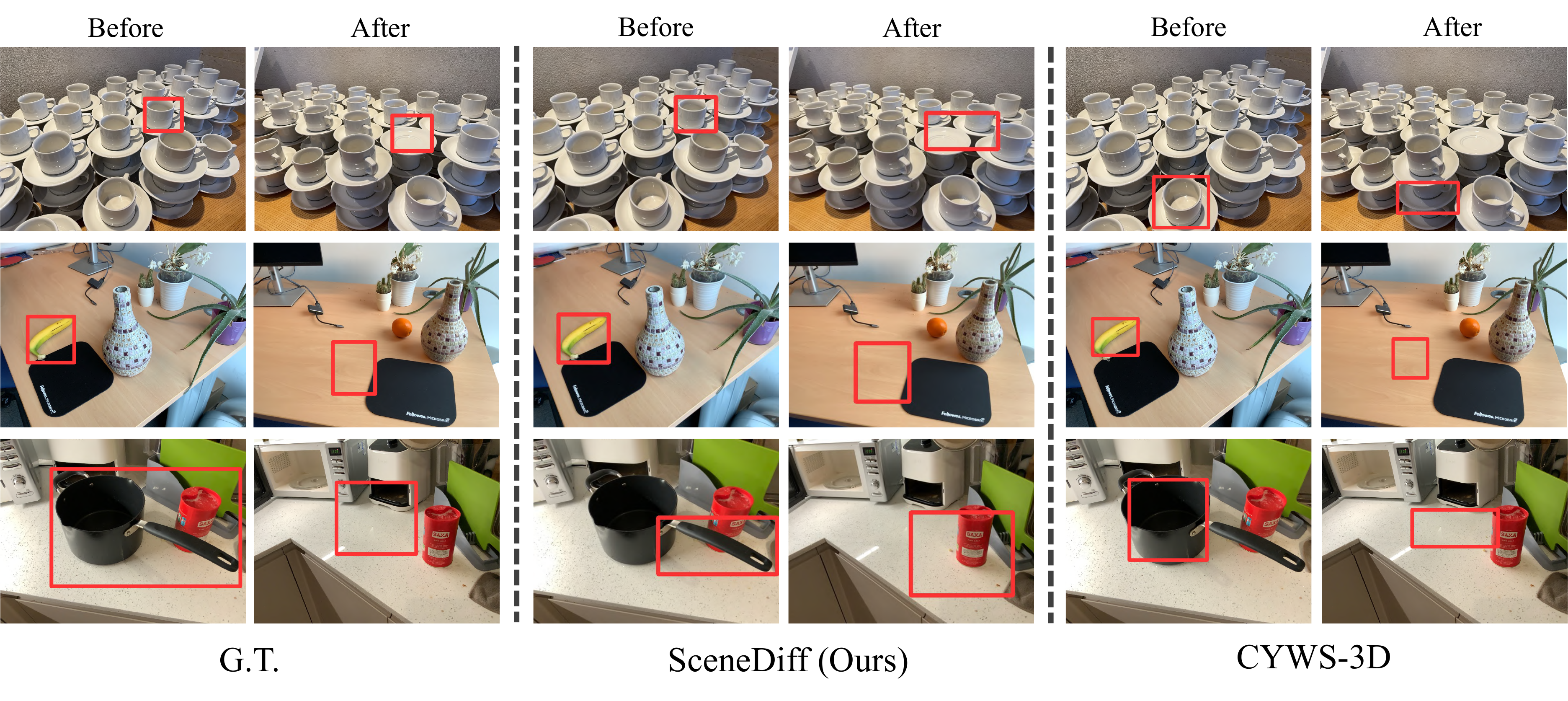}
\vspace{-2.4em}
\caption{
    \textbf{Qualitative Comparison on RC-3D~\cite{Sachdeva2023TheCY}.} 
    We visualize the highest-confidence detections for a single removed object per image pair for Ours and CYWS-3D~\cite{Sachdeva2023TheCY}. SceneDiff relies on SAM-generated masks for prediction, and can misalign with ground truth annotations (3rd row).
}
\vspace{-1em}
\label{fig:quualitative_rc3d}
\end{figure*}

\begin{table*}[h]
\centering
\caption{
\raggedright
\textbf{Ablation Study on SD-V test set.}
$E_{\{g,f,r\}}$ denote $E_{\{\text{geom}, \text{feat}, \text{region}\}}$.
Values show the difference ($\Delta$) from the full method ($\pi^3$+DINOv3).
}
\vspace{-0.8em}

\setlength{\tabcolsep}{3pt}
\renewcommand{\arraystretch}{1.05}

\resizebox{0.97\textwidth}{!}{
\begin{tabular}{c|ccccc|cccccc}
\toprule
\multirow{3}{*}{\textbf{Metric}}
 & \multicolumn{5}{c|}{\textbf{Backbone}} 
 & \multicolumn{6}{c}{\textbf{Feature Cues ($\pi^3$+DINOv3)}} \\
\cmidrule{2-12}
 & \multirow{2}{*}{\begin{tabular}[c]{@{}c@{}}$\pi^3$\\+DINOv3\end{tabular}}  & \multirow{2}{*}{\begin{tabular}[c]{@{}c@{}}$\pi^3$\\+DINOv2\end{tabular}} & \multirow{2}{*}{\begin{tabular}[c]{@{}c@{}}$\pi^3$\\+DINOv1\end{tabular}} & \multirow{2}{*}{\begin{tabular}[c]{@{}c@{}}VGGT\\+DINOv3\end{tabular}}  & \multirow{2}{*}{\begin{tabular}[c]{@{}c@{}}FASt3R\\+DINOv3\end{tabular}} & \multirow{2}{*}{$E_{g}$} & \multirow{2}{*}{$E_{f}$} & \multirow{2}{*}{$E_{r}$} & \multirow{2}{*}{\begin{tabular}[c]{@{}c@{}}$E_{g}$+$E_{f}$\end{tabular}} & \multirow{2}{*}{\begin{tabular}[c]{@{}c@{}}$E_{g}$+$E_{r}$\end{tabular}} & \multirow{2}{*}{\begin{tabular}[c]{@{}c@{}}$E_{f}$+$E_{r}$\end{tabular}} \\
 & & & & & & & & & \\
\midrule

\textit{px/im IoU}
& 39.1 & \textcolor{green!60!black}{+0.4} & \textcolor{red}{-3.0} & \textcolor{red}{-2.2} & \textcolor{red}{-23.7} 
& \textbf{\textcolor{green!60!black}{+2.1}} & \textcolor{red}{-6.1} & \textcolor{red}{-14.8} & \textcolor{red}{-1.8} & \textcolor{red}{-1.0} & \textcolor{red}{-4.5} \\
\textit{obj/im AP}
& 43.8 & \textcolor{green!60!black}{+1.3} & \textbf{\textcolor{green!60!black}{+1.5}} & \textcolor{red}{-8.1} & \textcolor{red}{-39.8}
& \textcolor{red}{-12.1} & \textcolor{green!60!black}{+0.1} & \textcolor{red}{-14.2} & \textcolor{red}{-0.1} & \textcolor{red}{-6.9} & \textcolor{green!60!black}{+0.9} \\
\textit{obj/sc AP}
& 22.8 & \textcolor{red}{-1.5} & \textcolor{red}{-2.7} & \textcolor{red}{-8.7} & \textcolor{red}{-22.4} 
& \textcolor{red}{-10.3} & 0.0 & \textcolor{red}{-0.8} & \textcolor{green!60!black}{+0.1} & \textcolor{red}{-5.0} & \textbf{\textcolor{green!60!black}{+2.6}} \\
\bottomrule
\end{tabular}
}

\vspace{-1.4em}
\label{tab:ablation}
\end{table*}
\vspace{-0.5em}
\subsection{Two-View Change Detection}
\vspace{-0.1em}
\label{sec:5.2}
We compare our method with existing methods on RC-3D~\cite{Sachdeva2023TheCY} in Tab.~\ref{tab:two_view_tab}, and show that SceneDiff outperforms all other methods.
Qualitative results are provided in Fig.~\ref{fig:quualitative_rc3d}. 
Although pixelwise change detection under similar viewpoints is not our primary target scenario, SceneDiff still outperforms the best existing work~\cite{Kim_2025_CVPR} by 2.2 points in F1 score on ChangeSim~\cite{park2021changesim} (see Supp. for details).

\vspace{-0.5em}
\subsection{Ablation Studies}
\vspace{-0.1em}
\label{sec:5.3}
We conduct experiments to evaluate our key design choices and the impact of different 3D estimation models~\cite{wang2025pi3, Wang2025VGGTVG, Yang2025Fast3RT3} and appearance feature extractors~\cite{simeoni2025dinov3, Oquab2023DINOv2LR, Caron2021EmergingPI} in Tab.~\ref{tab:ablation}.
Our experiments show that the geometry model really matters for alignment, as the reprojected appearance cue ($E_{\text{feat}}$) is the most important, significantly outperforming the region-level appearance matching cue ($E_{\text{region}}$) that does not require geometry reprojection. 
The geometry-based change features ($E_{\text{geom}}$) alone are effective for pixel segmentation but not very accurate for scene-level change detection. 
This may be due to the features being overly sensitive to geometric errors and incapable of finding changes in flat or very small objects.
DINO features ~\cite{Caron2021EmergingPI, Oquab2023DINOv2LR, simeoni2025dinov3} achieve similar performance, with DINOv3 performing the best for object-level evaluation, demonstrating stronger capacity for recognizing the same objects across different viewpoints and locations.
\begin{figure}[t]
\centering
    \includegraphics[width=.99\textwidth]{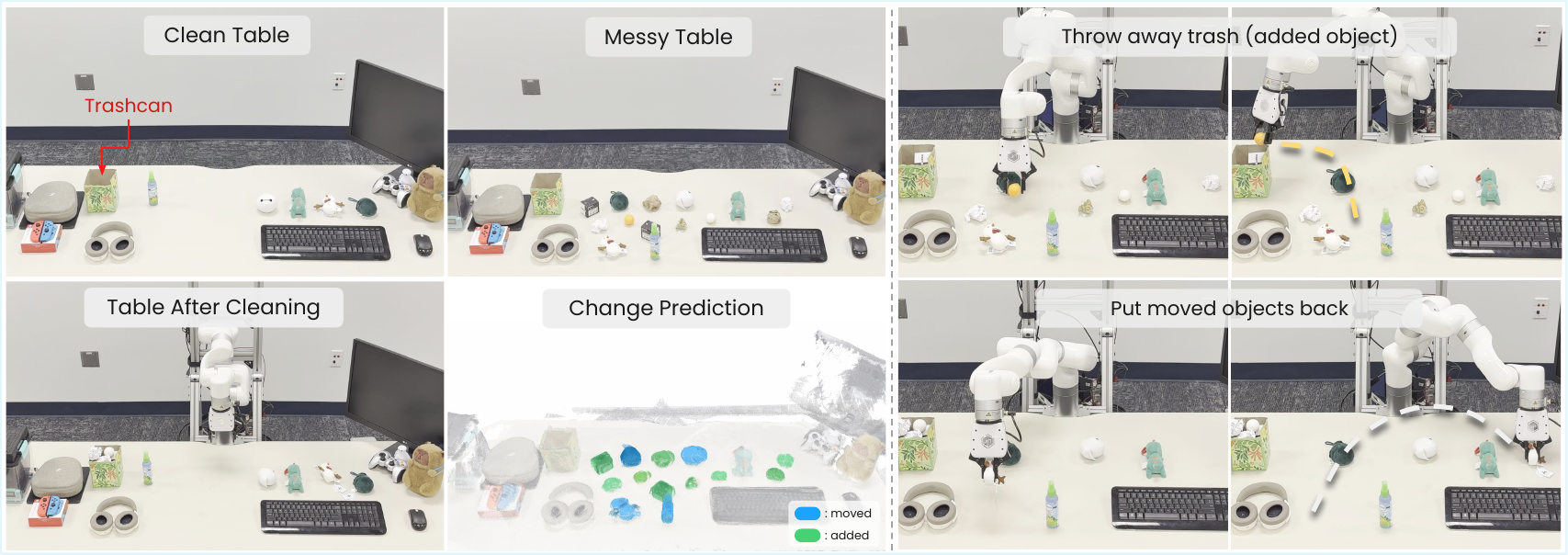}    
\vspace{-1.5mm}
\caption{
    \textbf{Real-World Cleaning Robot.} The left panel shows our initial "clean" and "messy" table state, and the right shows the robot's execution.
}
\label{fig:robot_demo}
\vspace{-.5em}
\end{figure}
\begin{figure*}[t]
\centering
\hspace{-0.5em}
        \includegraphics[width=0.98\textwidth]{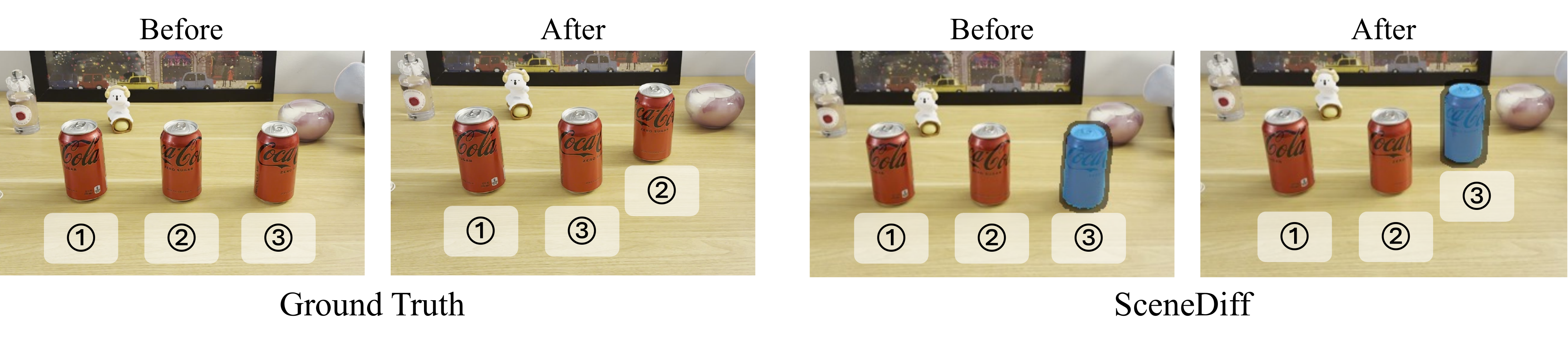}
\vspace{-1.em}
\caption{
    \textbf{Ambiguous Case.} When the second can is moved elsewhere and the third can is moved into its place, the observations are indistinguishable from only the third can having moved. SceneDiff therefore predicts only the third can as moved.
}
\label{fig:ambiguous}
\vspace{-1.em}
\end{figure*}

\vspace{-1.em}
\subsection{Applications}
\label{sec:robot_application}
\vspace{-0.2em}

SceneDiff enables various downstream tasks, especially in robotic manipulation. In Fig.~\ref{fig:robot_demo}, we show a demonstration of a robot cleaning a messy table back to the initial clean state by applying SceneDiff to detect moved and added objects. Please refer to Supp. for the video and more details and visualizations.

\vspace{-0.5em}
\section{Conclusion and Limitations}
\vspace{-0.5em}
\noindent
\hspace{-0.1em}
We presented the SceneDiff Benchmark, the first multiview dataset for object-level change detection across different camera trajectories, with 350 video pairs and dense instance-level annotations. Our training-free SceneDiff algorithm co-registers scenes using pretrained 3D, segmentation, and appearance models, then detects changes through geometric and semantic inconsistencies at the object level—generalizing across domains without retraining and naturally improving as foundation models advance. Experiments show large gains over prior work (53.0\% and 30.6\% relative AP on our benchmark and RC-3D, respectively), and we demonstrate practical utility through a closed-loop robotic tidying application. The benchmark, annotation tool, and code will be publicly released.


\vspace{1mm}
\noindent\textbf{Limitations.} 
Some change configurations are inherently ambiguous from geometry and appearance alone, e.g., swaps between visually identical objects (Fig.~\ref{fig:ambiguous}).
Furthermore, our method relies on pose estimation which can fail in some circumstances, such as extreme low-light, textureless surfaces, or sparse views with low co-visibility. Lastly, we focus on object-level spatial changes; semantic state changes (e.g., bread to toast) or fine-grained surface deformations (e.g., cracks) remain future work.

\noindent \textbf{Acknowledgement}
This work is supported in part by NSF IIS grant 2312102.
S.W. is supported by NSF 2331878 and 2340254, and research grants from Intel, Amazon, and IBM. 
This research used the Delta advanced computing resource, a joint effort of UIUC and NCSA supported by NSF (award OAC 2005572) and the State of Illinois.
Special thanks to Prachi Garg and Yunze Man for helpful discussion during project development, Bowei Chen, Zhen Zhu, Ansel Blume, Chang Liu, and Hongchi Xia for general advice and feedback on the paper, and Haoqing Wang and Vladimir Yesayan for data collection and annotation.

%
%
\bibliographystyle{splncs04}
\bibliography{main}

\clearpage
\begin{center}
{\Large{Supplementary Material}}
\end{center}

In the supplemental material, we first present additional results and analysis, including the cleaning robot demonstration~(Sec.~\ref{supp:robot_exp}), predictions on dynamic content~(Sec.~\ref{supp:dynamic_pred}), and results on ChangeSim~(Sec.~\ref{supp:changesim}) and PASLCD (Sec.~\ref{supp:PASLCD}). We then analyze failure cases~(Sec.~\ref{supp:failure_case}), hyperparameter sensitivity~(Sec.~\ref{supp:different_fixed_threshold}), geometry model robustness under varying change conditions~(Sec.~\ref{supp:geom_model_varying_change}), comparisons between different geometry models~(Sec.~\ref{supp:geom_model_comp}), additional qualitative results~(Sec.~\ref{supp:additional_qualitative}), and change-type-aware evaluation~(Sec.~\ref{change_type_aware_results}). 

Next, we detail our experimental setup, including device and running 
time ~(Sec.~\ref{supp:implement}), camera trajectory 
visualization~(Sec.~\ref{supp:camera_trajectory}), RC3D 
evaluation~(Sec.~\ref{supp:metric_rc3d}), and analysis and implementation details for 
baselines including 3DGS-based methods~(Sec.~\ref{supp:3DGS-CD}), 
VLMs (Sec.~\ref{supp:VLM}), CYWS-2D~(Sec.~\ref{supp:CYWS-2D}), and CYWS-3D~(Sec.~\ref{supp:CYWS-3D}). 

Finally, we describe the annotation pipeline~(Sec.~\ref{supp:annotation}) and 
data collection guidelines~(Sec.~\ref{supp:collection}).

Please see the included supplementary html file for videos of (1) our cleaning robot operation, (2) annotated examples, and (3) the annotation workflow for sequences. To preserve anonymity, the data will be publicly released upon acceptance. The self-contained codebase is attached in the supplementary zip file.
\section{Additional Results}
\label{supp:additional_results}

\subsection{Cleaning Robot}
\label{supp:robot_exp}
\begin{figure}[ht]
\centering
    \includegraphics[width=0.98\textwidth]{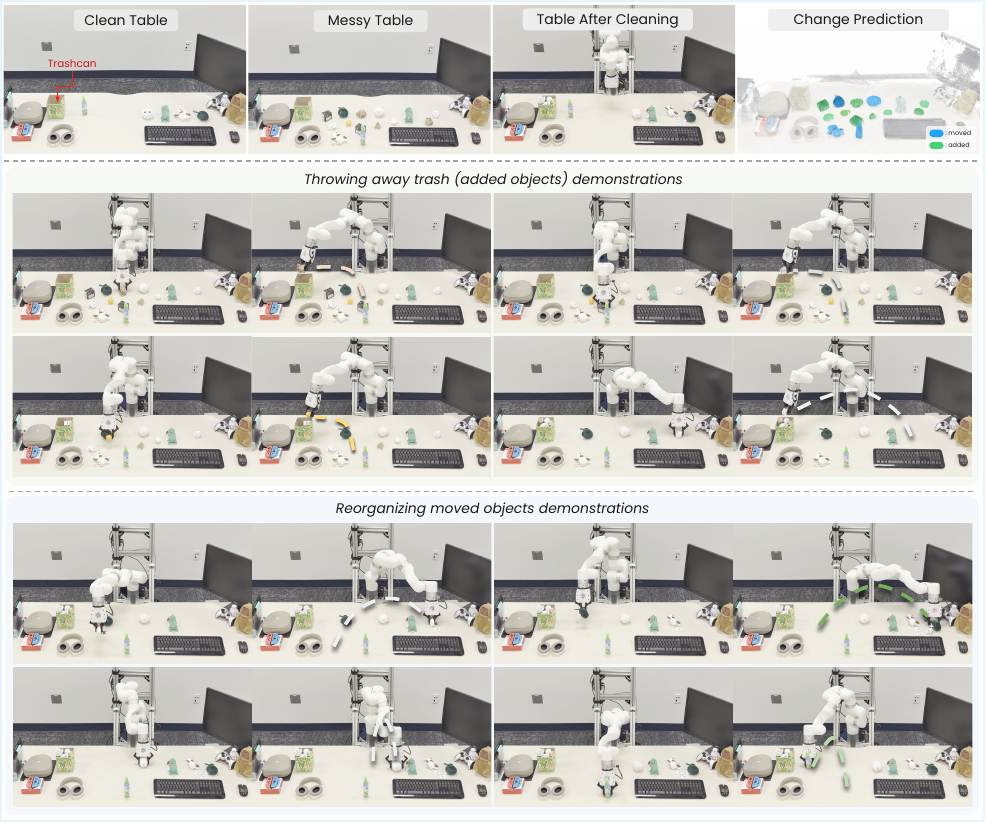}    
\vspace{-3mm}
\caption{
    \textbf{Cleaning Bot.} Top panel shows our "clean" and "messy" table state, the table after cleaning, and SceneDiff prediction. Subsequent panels show robots demonstration. 
}
\label{fig:supp_robot_demo}
\vspace{-1.5em}
\end{figure}

We additionally show a more detailed demonstration of our cleaning robot in  Fig.~\ref{fig:supp_robot_demo} and the supplementary video.

\vspace{-2mm}
\noindent \paragraph{Setup} We use the 7-dof UFACTORY xArm 7 robot arm for our demonstration. We first organize the table, setup a trashcan, and identify a few objects as \textit{moved} candidates. Then, we capture a video sequence of this \textit{before} setup. After that, we clutter the table with \textit{added} objects and randomly shift the \textit{moved} objects. Then, We capture this \textit{after} setup. 

\vspace{-2mm}
\noindent \paragraph{Coordinates Alignment} To better align $\pi^3$ prediction with the real-world coordinate, we capture one frame of the \textit{after} setup with a RealSense stereo camera and add it to the \textit{after} video sequence. We then run $\pi^3$ on the \textit{after} video sequence and treat the RealSense frame as the origin. After that, we align the $\pi^3$ prediction with the depth captured by the RealSense.

\vspace{-2mm}
\noindent \paragraph{Robot Movement} For each \textit{added} object predicted by SceneDiff, we compute the 3D centroid of the object as the grasp point. Similarly for \textit{moved} objects, we compute the before and after 3D centroids of the object. We use inverse kinematic to compute the robot joint angles given these centroids for movement. 

We additionally show a more detailed demonstration of our cleaning robot in  Fig.~\ref{fig:supp_robot_demo} and the supplementary video.

\clearpage
\subsection{Results with Dynamic Contents}
\label{supp:dynamic_pred}
\begin{figure*}[ht]
\vspace{-2em}
\centering
        \includegraphics[width=0.98\textwidth]{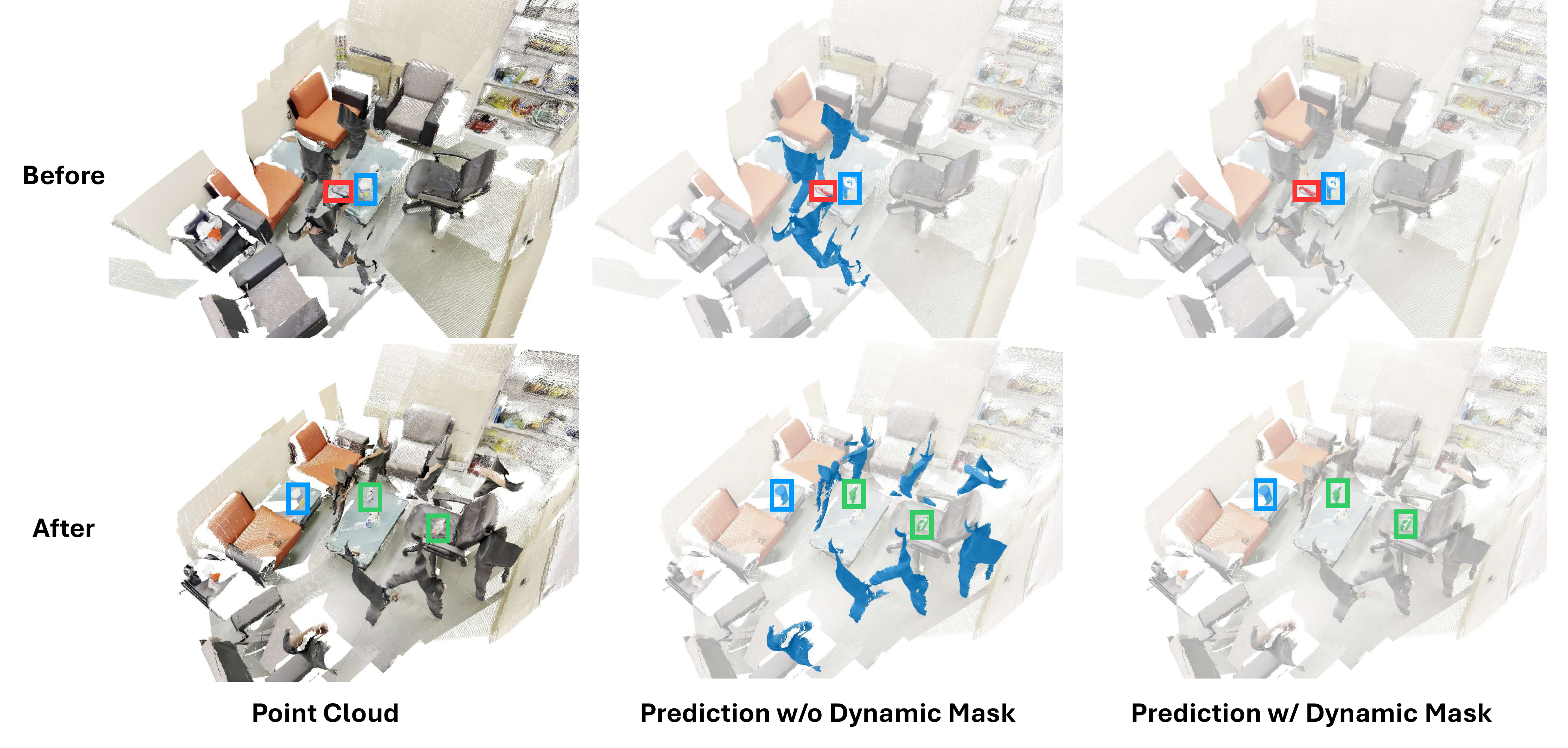}
\vspace{-1em}
\caption{
\textbf{Results Given Sequence Pair With Dynamic Content.}
From left to right, we show RGB point clouds, predictions from our method without dynamic masks, and predictions from our method with dynamic masks.
Changed objects are annotated using bounding boxes with the color scheme:
\textcolor[RGB]{255,51,51}{\texttt{Removed}}, 
\textcolor[RGB]{51,204,102}{\texttt{Added}}, and 
\textcolor[RGB]{0,153,255}{\texttt{Moved}}.
}
\vspace{-3em}
\label{fig:supp_dynamic_content}
\end{figure*}

We present our method's predictions on a sequence pair containing a moving person during capture in Fig.~\ref{fig:supp_dynamic_content}. 
To evaluate robustness to dynamic content, we manually annotate the person and provide this dynamic mask as an occlusion mask to our method.The results show that current geometry models can reconstruct the scene with the dynamic contents accurately, and without the dynamic mask, our method correctly predicts the changed objects but also identifies the person as a moved object in both sequences. When the dynamic mask is provided, our method successfully ignores the person and accurately predicts only the scene changes.

\subsection{Results on ChangeSim}
\label{supp:changesim}
Although our primary focus is on image pairs with diverse viewpoints, we provide comparisons with existing work~\cite{Kim_2025_CVPR,wang2023reduce,chen2021dr,liu2021super,sakurada2020weakly} on ChangeSim~\cite{park2021changesim} in Table~\ref{tab:two_view_similar_pose_tab}. 
We generate the output using the two-view input setting (Sec.~4.4), project the predicted change from the reference (training) view onto the query (test) view, and take the union with the predicted change in the query view to achieve a symmetric change prediction.
Despite ChangeSim's similar-viewpoint setting, our method still outperforms existing methods.
\begin{table}[h]
\centering
\renewcommand{\arraystretch}{1.2}
\vspace{-1em}
\caption{
\textbf{Two-View Change Detection on Changesim~\cite{park2021changesim}.}
F1-scores for pixel-level change segmentation are reported.
Results for other methods are from the original papers.
}
\resizebox{0.7\textwidth}{!}{
\begin{tabular}{l@{\hspace{6pt}}|@{\hspace{6pt}}c@{\hspace{6pt}}c@{\hspace{6pt}}c@{\hspace{6pt}}c@{\hspace{6pt}}c@{\hspace{6pt}}c}
\toprule
Method & \textbf{SceneDiff} & GeSCF & C-3PO & DR-TANet & CDResNet &  CSCDNet \\ 
\midrule
\textit{F1} & \textbf{57.0} & 54.8 & 44.7 & 40.3 & 41.3 & 43.1  \\ 
\bottomrule
\end{tabular}
}
\label{tab:two_view_similar_pose_tab}
\end{table}


\subsection{Results on PASLCD}
\label{supp:PASLCD}
\begin{figure*}[h]
\vspace{-2em}
\centering
        \includegraphics[width=0.98\textwidth]{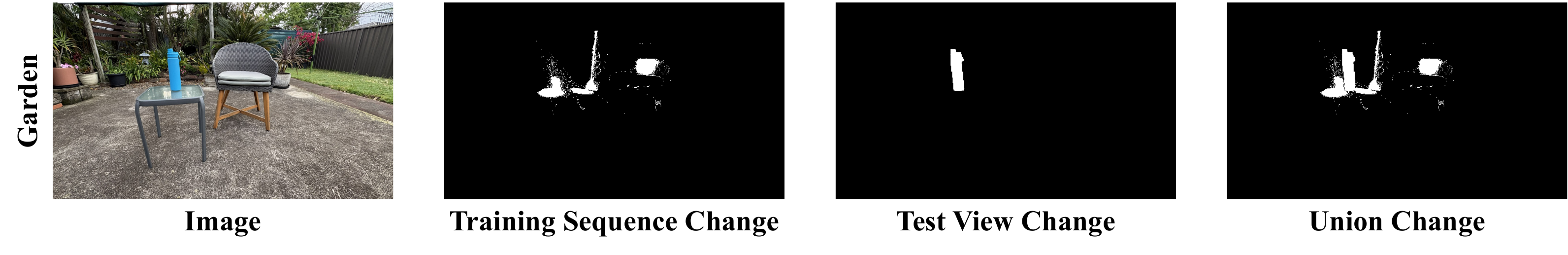}
\vspace{-1em}
\caption{
\textbf{Symmetric Change Example.} We visualize the input image, change reprojected from the training sequence, change in test view, and the final union change.
}
\vspace{-2.2em}
\label{fig:supp_symmetric}
\end{figure*}

\begin{figure*}[h]
\vspace{-2em}
\centering
        \includegraphics[width=0.98\textwidth]{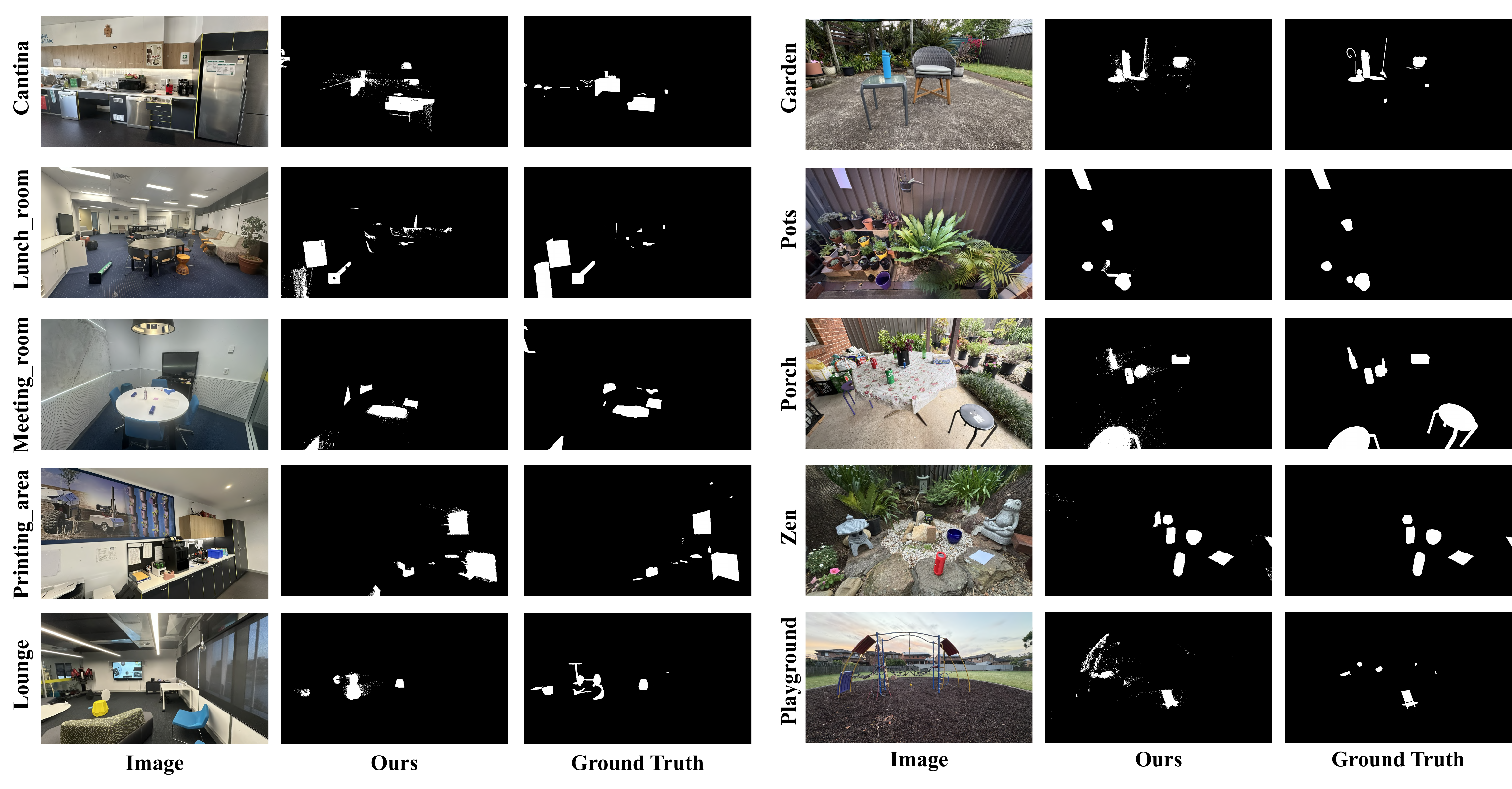}
\vspace{-1.2em}
\caption{
\textbf{Qualitative Results on PASLCD.} We visualize input images, predicted symmetric change, and ground truth from the test view of each scene . 
}
\vspace{-4.2em}
\label{fig:supp_paslcd}
\end{figure*}

We evaluate our method on PASLCD~\cite{Galappaththige2024MultiViewPC}. 
We run the standard pipeline to predict the object-level change in both sequences, and project the predicted change from the training (before change) sequence onto the test (after change) sequence, and take the union with the predicted change in the test sequence to achieve a symmetric change prediction (Fig.~\ref{fig:supp_symmetric}).
See Fig.~\ref{fig:supp_paslcd} for qualitative results and Tab.~\ref{tab:supp_paslcd} for full quantitative results.

\begin{table*}[h]
    \centering
    \caption{\textbf{Quantitative results on PASLCD.} We report the mIoU values averaged across similar and different lighting condition instances of both Indoor and Outdoor scenes. Baseline numbers are sourced from MV3DCD~\cite{Galappaththige2024MultiViewPC}.}
    \vspace{-0.8em}
    \scalebox{0.7}{
        \begin{tabular}{l@{\hspace{6pt}}c@{\hspace{6pt}}c@{\hspace{6pt}}c@{\hspace{6pt}}c@{\hspace{6pt}}c@{\hspace{6pt}}c}
            \toprule
            Scene & FF/360 & SplatPose~\cite{Kruse2024SplatPoseD} & CSCDNet~\cite{sakurada2020weakly} & CYWS-2D~\cite{sachdeva2023change} & MV3DCD~\cite{Galappaththige2024MultiViewPC} & Ours \\
            \midrule
            Cantina & FF & 0.188 & 0.079 & 0.277 & \textbf{0.580} & 0.473 \\
            Lounge & FF & 0.262 & 0.195 & 0.221 & \textbf{0.463} & 0.367 \\
            Printing Area & FF & 0.183 & 0.147 & 0.327 & 0.588 & \textbf{0.723} \\
            Lunch Room & 360 & 0.133 & 0.035 & 0.123 & \textbf{0.389} & 0.193 \\
            Meeting Room & 360 & 0.130 & 0.213 & 0.138 & 0.350 & \textbf{0.570} \\
            \midrule
            Garden & FF & 0.185 & 0.241 & 0.346 & 0.436 & \textbf{0.473} \\
            Pots & FF & 0.140 & 0.022 & 0.351 & \textbf{0.540} & 0.502 \\
            Zen & FF & 0.186 & 0.010 & 0.450 & 0.500 & \textbf{0.760} \\
            Playground & 360 & 0.081 & 0.134 & 0.059 & 0.249 & \textbf{0.258} \\
            Porch & 360 & 0.239 & 0.176 & 0.439 & \textbf{0.518} & 0.408 \\
            \midrule
            Average & -- & 0.173 & 0.125 & 0.273 & 0.461 & \textbf{0.473} \\
            \bottomrule
        \end{tabular}
    }
\label{tab:supp_paslcd}
\end{table*}

\subsection{Failure Cases and Challenging Scenarios}
\label{supp:failure_case}
We analyze representative failure cases and challenging scenarios to understand the fundamental limitations of multiview change detection. 
We identify two primary failure modes where the method produces unreliable predictions: (1) ambiguity in large cluttered scenes with repetitive items (11 of 200 sequence pairs in SD-V), and (2) geometry reconstruction failure (2 of 200).
We also discuss strong lighting changes as a challenging scenario that degrades performance.

\begin{figure}[h]
\hspace{-4em}
\vspace{-1.5em}
\centering
\begin{tabular}{c}
        \includegraphics[width=0.74\linewidth]{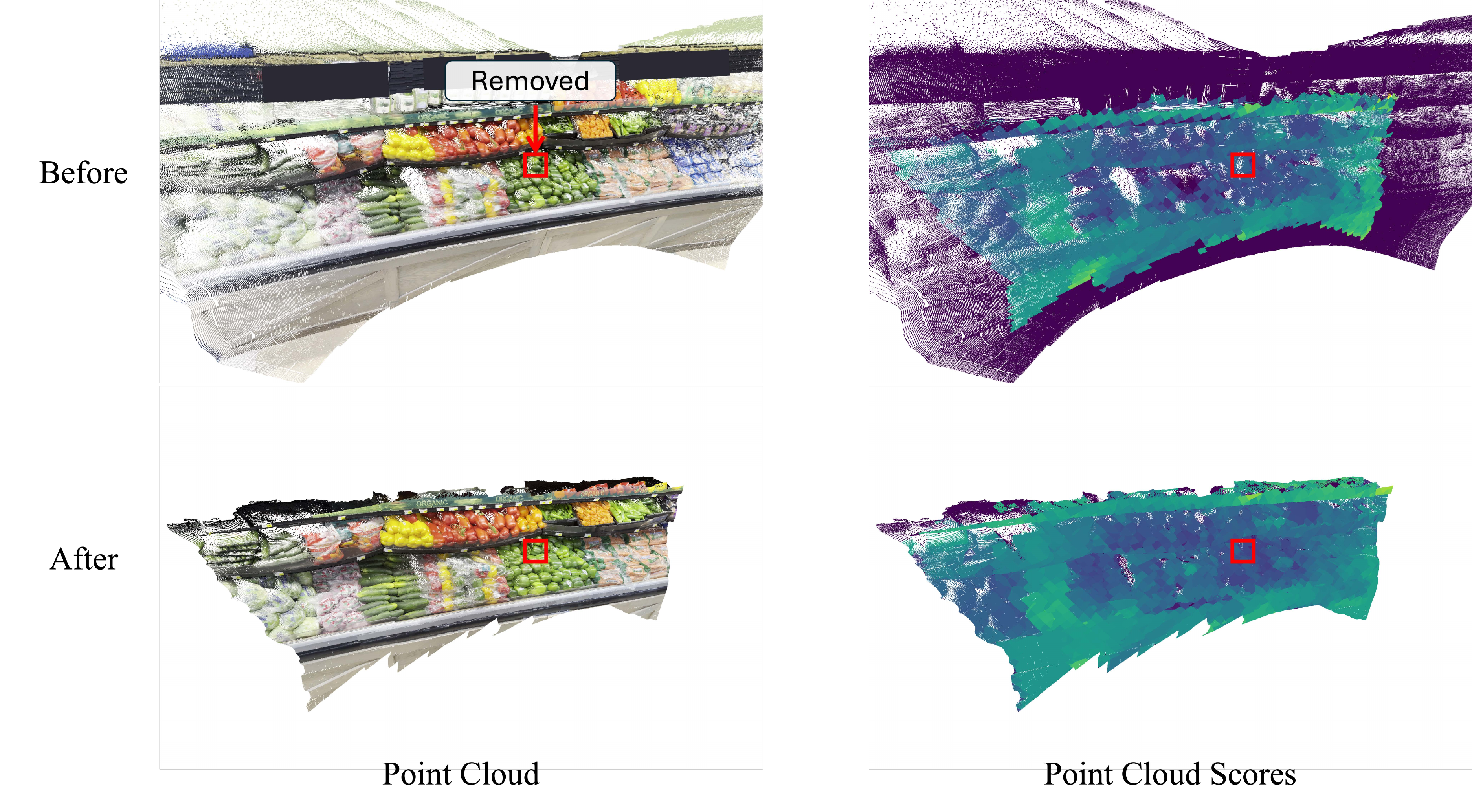}
\end{tabular}
\caption{
\textbf{Failure Analysis: Large Scenes With Cluttered Repetitive Items.}
We show the point clouds and corresponding change scores for a failure case 
in a large market scene with repetitive items. The removed object is marked 
with red boxes in both before and after point clouds. Despite the geometric change, the high density of similar items makes detection challenging (all metrics values at 0 in this scene). 
}
\label{fig:supp_repeatative_failure_analysis}
\vspace{-0.9em}
\end{figure}

\noindent \textbf{Large Cluttered Scenes with Repetitive Items.}
Large cluttered scenes with repetitive items (e.g., market shelves with 
identical products) present fundamental challenges for appearance-based 
change detection methods and represent the primary cause of failure in our 
dataset (11 cases).
Changed objects often have visually similar counterparts elsewhere in the scene, making the region-matching score $E_{\text{region}}$ unreliable, as it yields high appearance similarity regardless of actual changes. Additionally, when a foreground object is removed, the newly exposed background often contains 
similar-looking items, making the reprojected feature score 
$\mathbf{E}_{\text{feat}}$ less discriminative. In such cases, only the geometry reprojection score $\mathbf{E}_{\text{geom}}$ provides reliable change signals, which may be insufficient in large, densely packed scenes. Fig.~\ref{fig:supp_repeatative_failure_analysis} shows one representative 
failure case.

\begin{figure}[t]
\centering
\begin{tabular}{c}
        \includegraphics[width=0.9\linewidth]{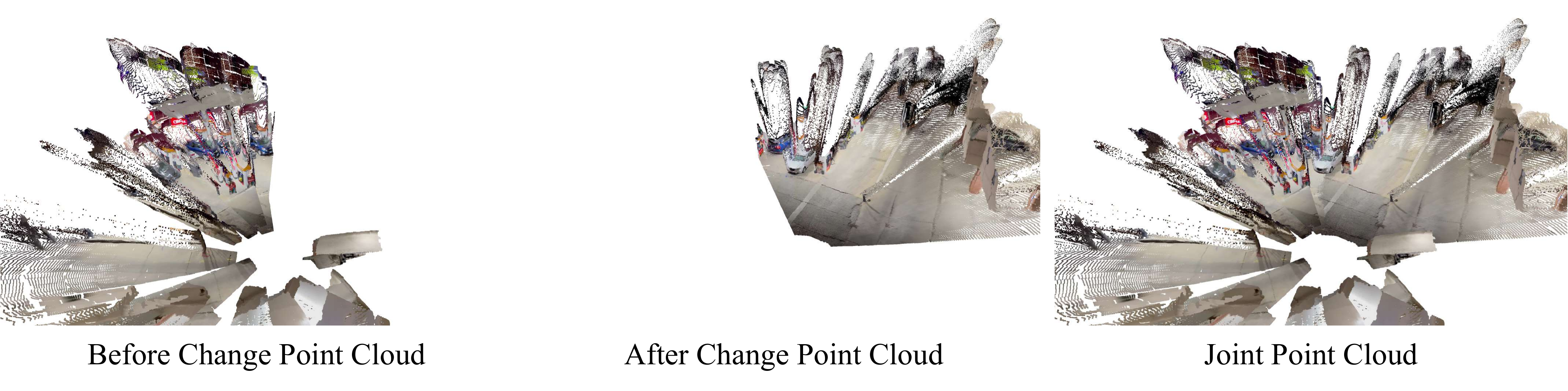}
\end{tabular}
\caption{
\textbf{Failure Analysis: Geometry Reconstruction Failure Due to Limited Overlap.} 
We show a failure case where the input sequence pair has limited overlap, 
as they view the gas station from significantly different angles. The three 
point cloud visualizations show the sequence before the change, the sequence 
after the change, and both sequences combined. The inaccurate geometry 
reconstruction causes complete detection failure (all metrics values at 0). 
Note that limited overlap between views is a well-known challenge in 
multi-view geometry reconstruction.
}
\vspace{-1em}
\label{fig:supp_geo_failure_analysis}
\end{figure}

\noindent \textbf{Geometry Failure.} 
Since our method relies on geometric models, reconstruction failures prevent accurate change detection, a challenge shared by all geometry-based approaches. 
For example, as shown in Fig.~\ref{fig:supp_geo_failure_analysis}, when the 
input sequence pair has limited overlap, joint geometry reconstruction can fail.
However, in our dataset, we do not observe geometry failures caused by the 
changes themselves between sequences, unless under stress testing 
(see Sec.~\ref{supp:geom_model_varying_change}).

\begin{figure*}
\hspace{-4em}
\centering
\begin{tabular}{c}
        \includegraphics[width=0.9\linewidth]{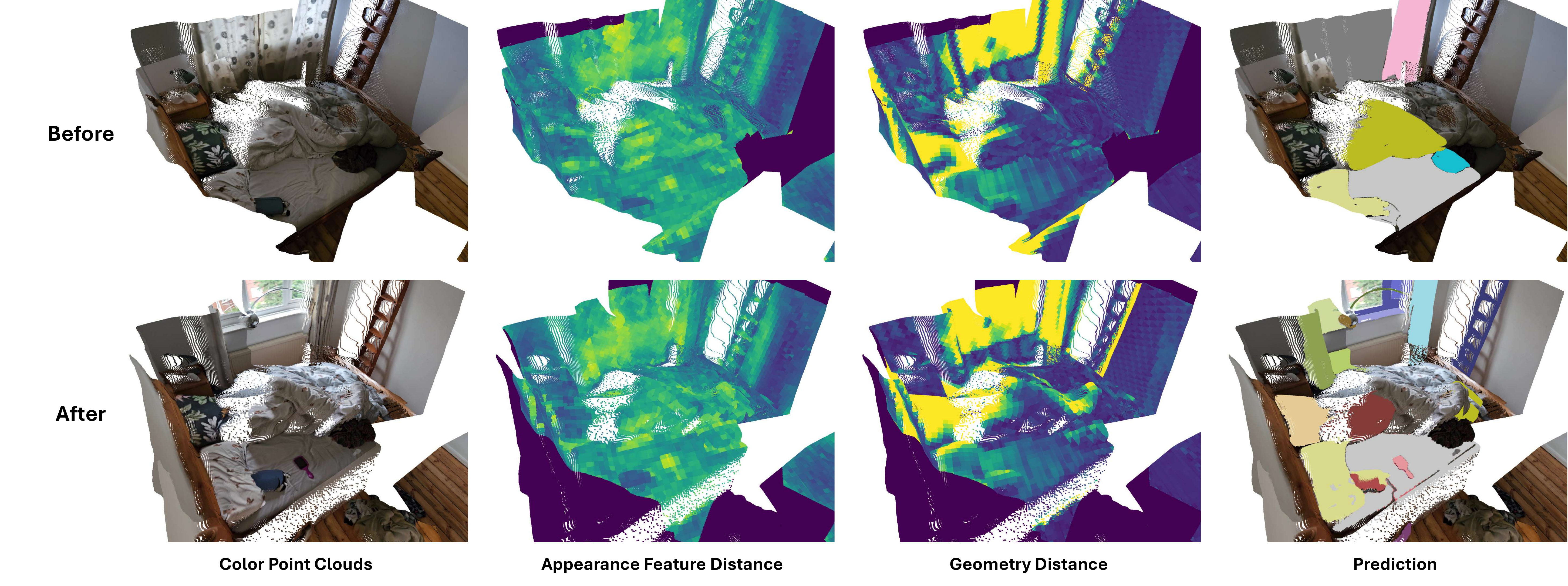}
\end{tabular}
\hspace{-4em}
\caption{
\textbf{Challenging Scenario: Sequence Pair with Strong Lighting Changes.} 
For each point in one sequence, we find the nearest point in the other sequence and compute the appearance feature distance (2nd column) and geometry distance (3rd column). Distances are scaled for visualization only and are not directly used by our method. Under strong lighting changes in large-scale scenes, geometry alignment degrades and unchanged objects exhibit different appearance features, resulting in increased false positive predictions (e.g., pillows and quilt marked as changed).}
\label{fig:supp_light_change_failure_analysis}
\end{figure*}

\noindent \textbf{Strong Lighting Changes.}
While not causing complete failure, strong lighting changes present a 
significant challenge that degrades detection accuracy.
As shown in Fig.~\ref{fig:supp_light_change_failure_analysis}, strong lighting changes degrade geometry reconstruction quality and affect appearance matching for static content, causing more false positive predictions. 
This is also a fundamental challenge for appearance-based methods, as distinguishing lighting variations from actual changes remains difficult. Note that lighting variations are more severe in the actual videos than apparent in the rendered point cloud visualizations.

\clearpage
\subsection{Hyperparameter Sensitivity Analysis}
\label{supp:different_fixed_threshold}
\begin{figure*}[ht]
\centering
\vspace{-1.5em}
        \includegraphics[width=0.98\textwidth]{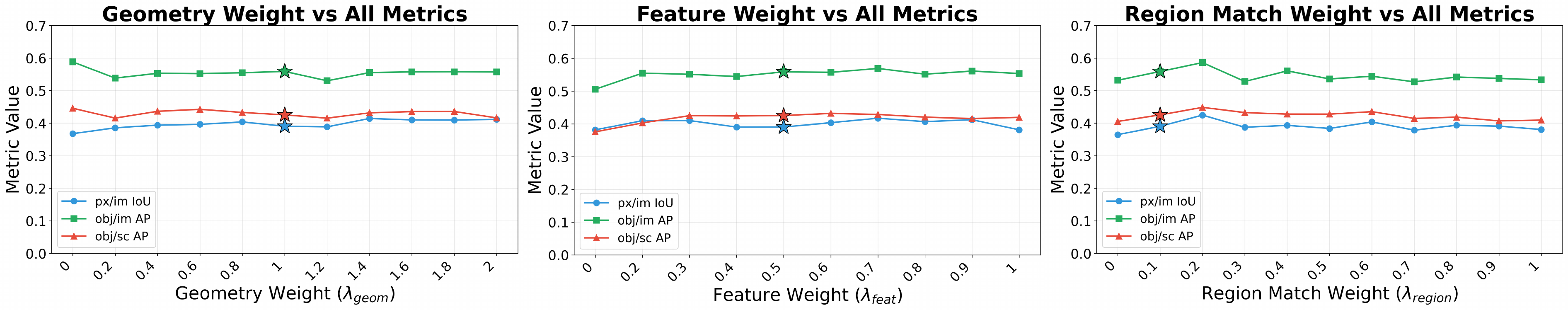}
\vspace{-1em}
\caption{
\textbf{Robustness Analysis for Weights.} 
We report \textit{px/im IoU}, \textit{obj/im AP}, and \textit{obj/sc AP} 
across varying geometry, feature, and region-level matching weights. Stars denote the default parameter values used in the main paper. The method demonstrates highly stable performance across a broad range of weights.
}
\label{fig:supp_robustness_weights}
\end{figure*}

\begin{figure*}[ht]
\centering
\hspace{-1em}
        \includegraphics[width=0.85\textwidth]{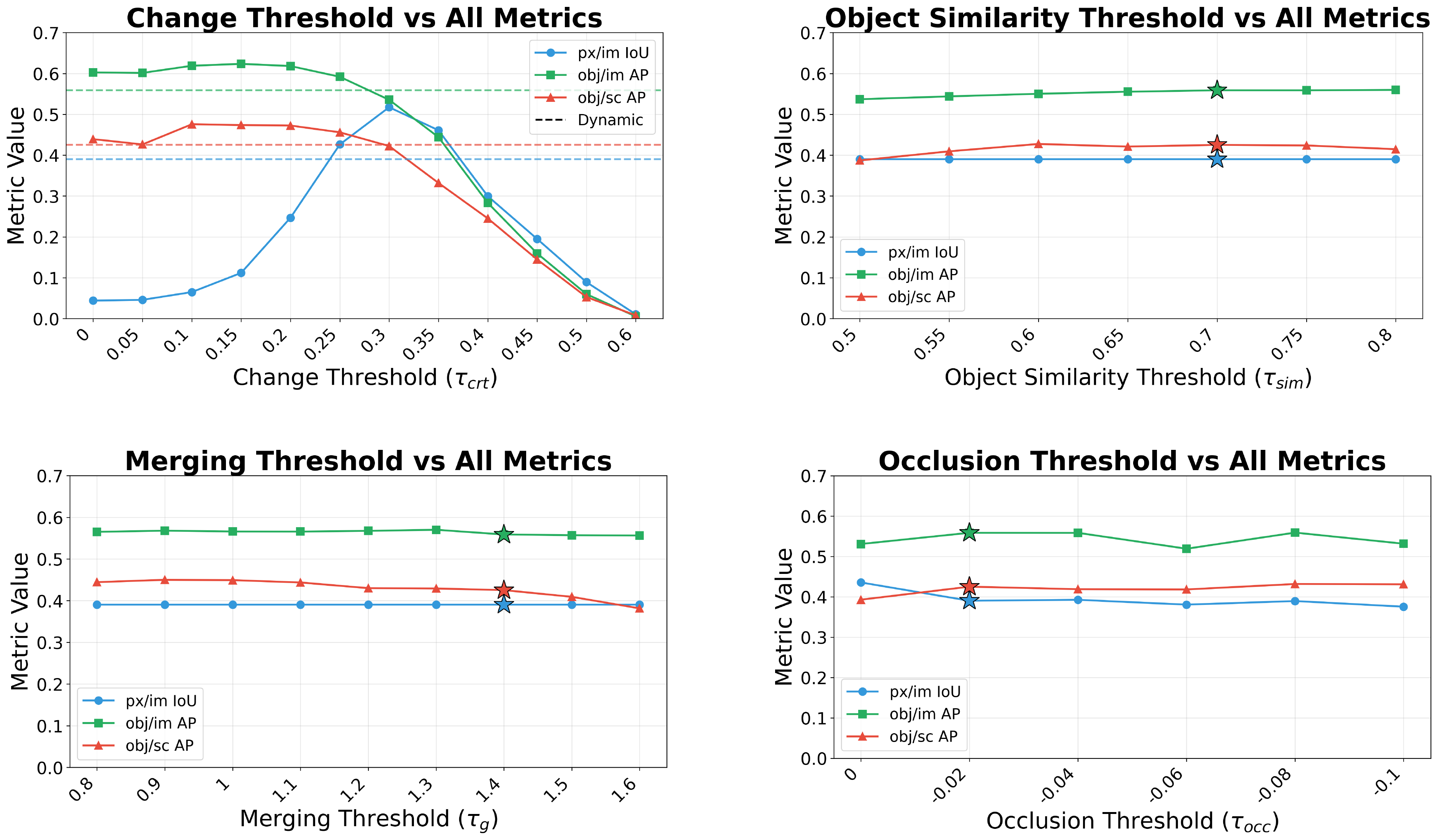}
\vspace{-1em}
\caption{
\textbf{Robustness Analysis for Thresholds.} 
We report \textit{px/im IoU}, \textit{obj/im AP}, and \textit{obj/sc AP} 
under varying change, merging, similarity, and occlusion thresholds. The dashed line represents the performance of our dynamic threshold, while stars indicate the fixed parameter values referenced in the main paper. The model exhibits low sensitivity to threshold variations.
}
\vspace{-1.5em}
\label{fig:supp_robustness_threshold}
\end{figure*}

We visualize performance comparisons on the SceneDiff validation set with varying hyperparameters --- geometry weights ($\tau_{geom}$), feature weights ($\tau_{feat}$), and region-level matching weights ($\tau_{region}$) in Fig.~\ref{fig:supp_robustness_weights}, and change thresholds ($\tau_{\Delta}$), merging thresholds ($\tau_{merge}$), object similarity thresholds ($\tau_{sim}$), occlusion thresholds ($\tau_{occ}$) in Fig.~\ref{fig:supp_robustness_threshold}. 
Notably, our method exhibits very low sensitivity to these hyperparameter choices, maintaining highly stable performance across a broad spectrum of values. While manually tuning for the optimal fixed change threshold provides a boost to the mIoU, the dynamic approach we employ (using max entropy thresholding) achieves highly competitive results and does not need manual tuning.

\vspace{-3em} 

\clearpage
\subsection{Geometry Models Under Varying Change Conditions}
\label{supp:geom_model_varying_change}
\begin{figure}[ht]
\vspace{-1em}
\centering
        \includegraphics[width=0.75\textwidth]{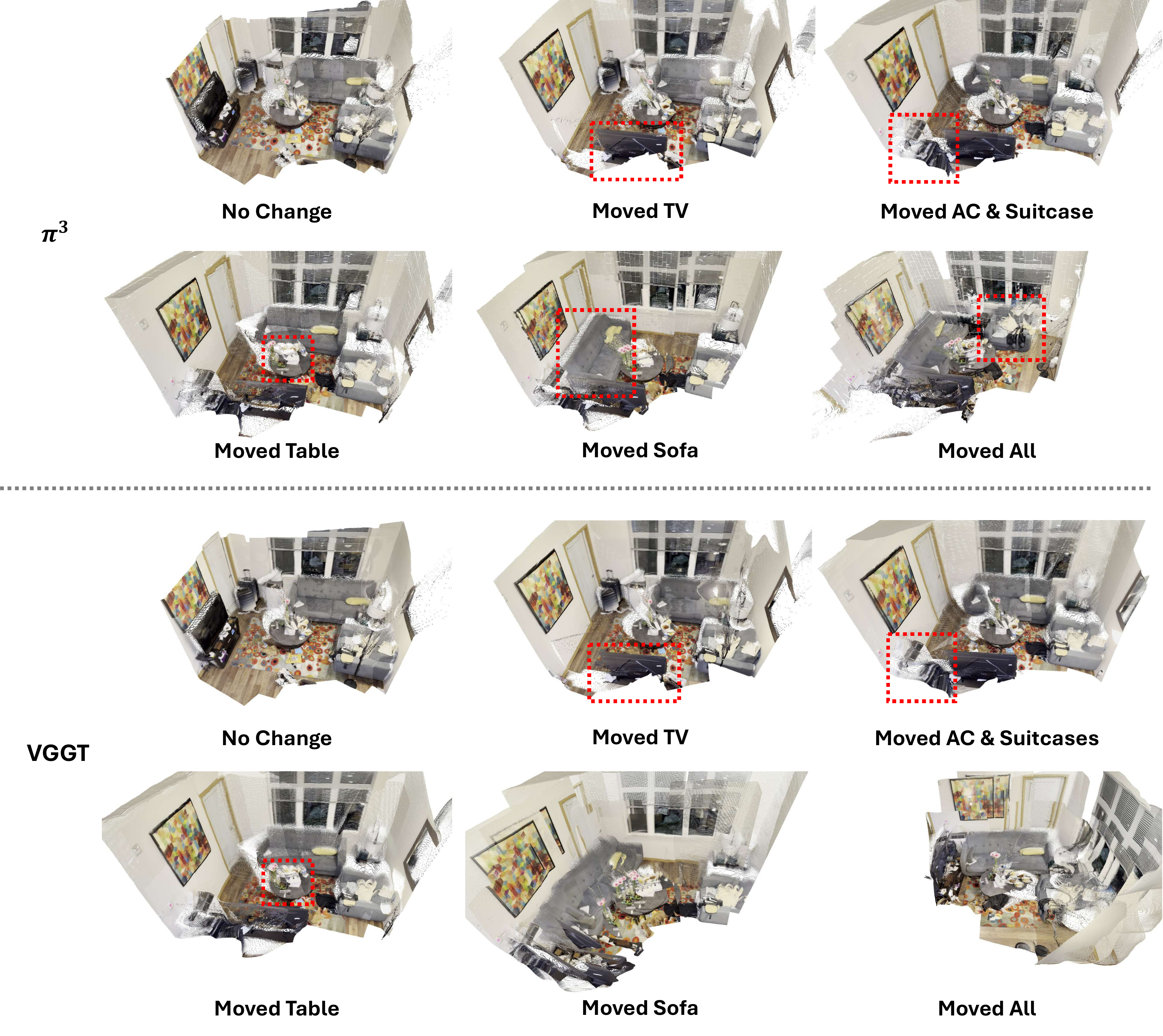}
\vspace{-1em}
\caption{
    \textbf{Geometry Models Under Varying Amount of Change.}
    To test the robustness of geometry models, we sequentially move the objects following similar transformations, e.g., the TV, AC\&Suitcase, Table, 1st Sofa, 2nd Sofa. 
    For each geometry model, we visualize the estimated point clouds of no-change sequence, and the sequence of each change state when fed together with the no-change sequence. 
    Point clouds are aligned and all rendered views come from the same camera pose. 
    Change objects (if recognizable) are marked with red boxes. 
}
\label{fig:geometry_models_different_change}
\end{figure}

We visualize geometry models' predictions under varying amounts of change in Fig.~\ref{fig:geometry_models_different_change}. Geometry models are generally robust to changes unless most objects undergo similar transformations simultaneously.

The figure shows that $\pi^3$~\cite{wang2025pi3} begins to produce less accurate geometry when all movable objects are changed following similar transformations, but can still align the two sequences reasonably. 
Interestingly, VGGT~\cite{Wang2025VGGTVG} fails to reconstruct point clouds when most movable objects are changed. When all movable objects are changed, it identifies the actual moved objects as static and the actual static contents (walls, windows) as moved, i.e., inverting the scene dynamics.
This behavior may arise from VGGT regressing point clouds under the first frame's coordinate system, causing the post-change sequence to always align with the first frame but not first align within the changed sequence. 
In contrast, $\pi^3$ regresses point clouds under an arbitrary coordinate system, allowing frames in the post-change sequence to first align among themselves before aligning with the pre-change sequence, thus exhibiting more robustness to change.

\subsection{Comparisons Between Geometry Models}
\label{supp:geom_model_comp}
\begin{figure*}[h]
\centering
\begin{tabular}{c}
        \includegraphics[width=0.7\linewidth]{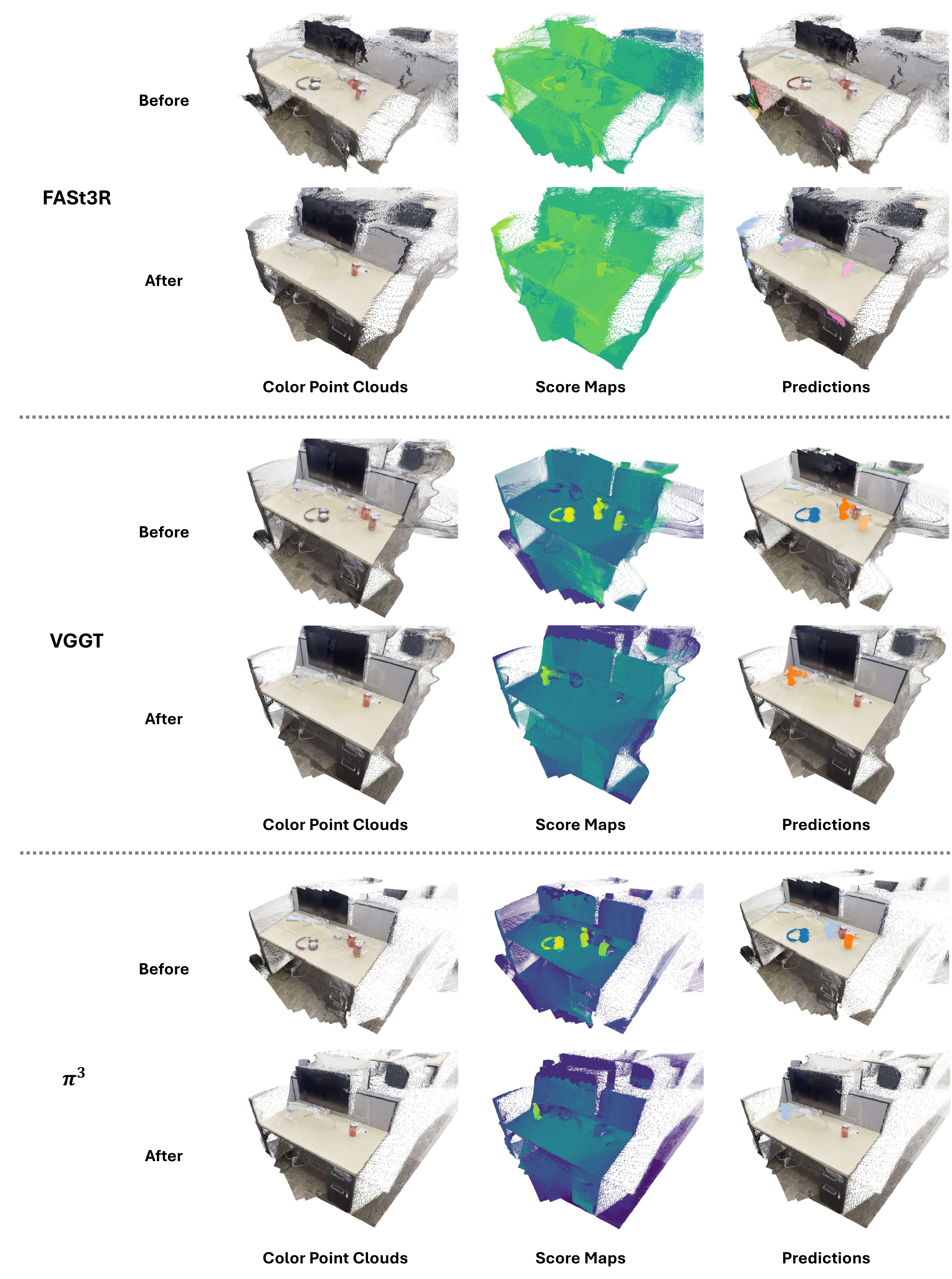}
\end{tabular}

\caption{
\textbf{Comparisons when using FASt3R, VGGT, and $\boldsymbol{\pi}^3$.} VGGT and $\pi^3$ produce precisely aligned point clouds across sequences, enabling accurate score maps through reliable geometry reprojection. In contrast, while FASt3R achieves reasonable alignment, it generates much noisier point clouds that degrade score map quality.
}
\label{fig:supp_geometry_model_vis}
\end{figure*}

We present comparisons using different geometry models~\cite{Wang2025VGGTVG, Yang2025Fast3RT3, wang2025pi3} in Fig.~\ref{fig:supp_geometry_model_vis}.

\subsection{Additional Qualitative Comparisons}
\label{supp:additional_qualitative}
\begin{figure}[h]
\centering
\begin{tabular}{c}
        \includegraphics[width=0.8\linewidth]{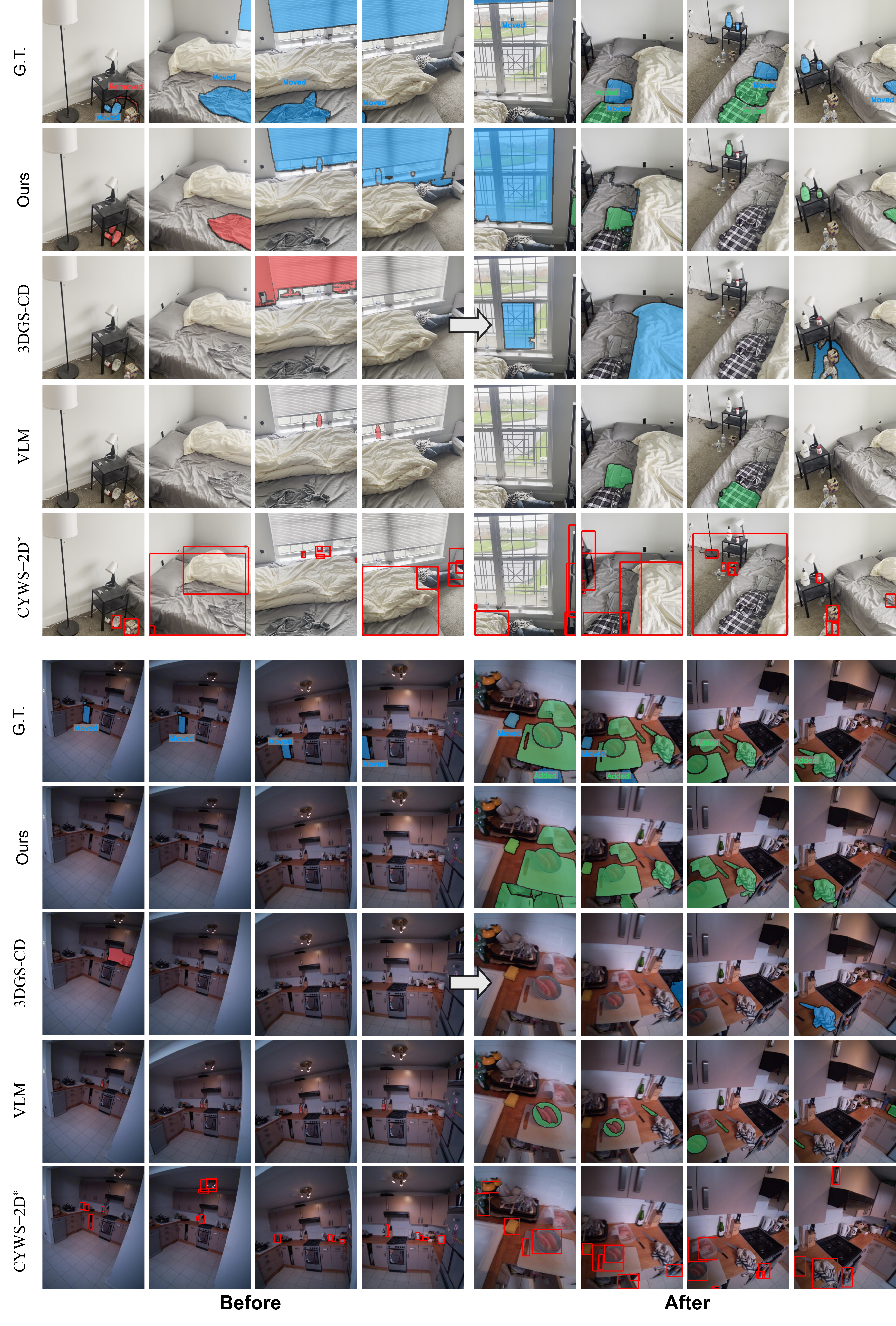}
\end{tabular}

\caption{
\raggedright
    \textbf{Comparisons in Challenging Sequence Pairs.} We compare our method (SceneDiff) with existing methods on more challenging sequence pairs. CYWS-2D is fine-tuned and visualized using the top 5 predictions, or the total number of ground truth objects in the view if that count exceeds 5. Color map: \textcolor[RGB]{255,51,51}{\texttt{Removed}}, \textcolor[RGB]{51,204,102}{\texttt{Added}}, and \textcolor[RGB]{0,153,255}{\texttt{Moved}}.
}
\label{fig:supp_comp}
\end{figure}

Additional comparisons with existing methods~\cite{sachdeva2023change, Lu20243DGSCD3G, Qwen2.5-VL} in Fig.~\ref{fig:supp_comp}.

\clearpage

\subsection{\textit{obj/sc AP} with Change Type Awareness}
\label{change_type_aware_results}
In the main paper, we evaluate scene-level object change detection (\textit{obj/sc AP}) using a change-type agnostic criterion. Specifically, detections across views that correspond to the same object hypothesis are aggregated into a single scene-level prediction, with the overall confidence computed as the mean of the per-view confidences. A prediction is then matched to a ground-truth object if the aggregated IoU across all frames in both sequences exceeds 0.5. 

To provide a more comprehensive analysis, we also report the \textit{obj/sc AP} under a stricter, change-type aware setting. In this formulation, a prediction is considered a true positive only if it both exceeds the 0.5 IoU threshold and correctly predicts the change type (i.e., added, removed, or moved). 

\begin{table*}[h]
\centering
\renewcommand{\arraystretch}{1.2}
\caption{
\textbf{\textit{obj/sc AP} Under Different Criteria.}
We provide both the evaluation in the main paper and the stricter change-type aware metric.
}
\resizebox{0.8\textwidth}{!}{
\begin{tabular}{l@{\hspace{6pt}}|c@{\hspace{6pt}}c@{\hspace{6pt}}c@{\hspace{6pt}}|c@{\hspace{6pt}}c@{\hspace{6pt}}c}
\toprule
& \multicolumn{3}{c|}{\textbf{SD-V}} & \multicolumn{3}{c}{\textbf{SD-K}}\\
 & 3DGS-CD & VLM  & SceneDiff & 3DGS-CD & VLM & SceneDiff \\ 
\midrule
Change-type agnostic & 0.5 & 5.3  & \textbf{22.8} & 0.1 & 2.1 & \textbf{10.6}  \\ 
Change-type aware & 0.3 & 2.5 & \textbf{18.5} & 0.1 & 1.5 & \textbf{7.7} \\ 
\bottomrule
\end{tabular}
}
\label{tab:supp_obj_sc_ap}
\end{table*}

\clearpage

\section{Experimental Details}
\label{supp:exp_details}




\subsection{Device and Running Time}
\label{supp:implement}
We run all experiments with a single NVIDIA A40. The average running time of each method is provided in Tab.~\ref{tab:supp_time}. 
The inference time of 3DGS-CD and MV3DCD include the training time of 3D Gaussian Splatting. The major computational cost of our method comes from generating masks for each input view, taking around 4 seconds per image since we generate both \textit{whole} and \textit{part} masks from SAM~\cite{Kirillov2023SegmentA} and derive non-overlapping instance masks as regions from them.


\begin{table*}[h]
\caption{
    \textbf{Average Inference Time of A Sequence Pair on SceneDiff Benchmark.} 
}
\centering
\resizebox{0.7\textwidth}{!}{
\begin{tabular}{l@{\hspace{6pt}}|c@{\hspace{6pt}}c@{\hspace{6pt}}c@{\hspace{6pt}}c@{\hspace{6pt}}c@{\hspace{6pt}}c}
\toprule
Method & SceneDiff & VLM  & CYWS-2D & CYWS-3D & 3DGS-CD & MV3DCD \\
\midrule
Time & 159.4s & 30.5s & 18.4s & 23.9s & >30mins & >30mins \\  
\bottomrule
\end{tabular}
}
\vspace{-2em}
\label{tab:supp_time}
\end{table*}

\subsection{Camera Trajectory Visualization}
\label{supp:camera_trajectory}
\begin{figure}[h]
\vspace{-2em}
\centering
\begin{tabular}{c}
        \includegraphics[width=0.85\linewidth]{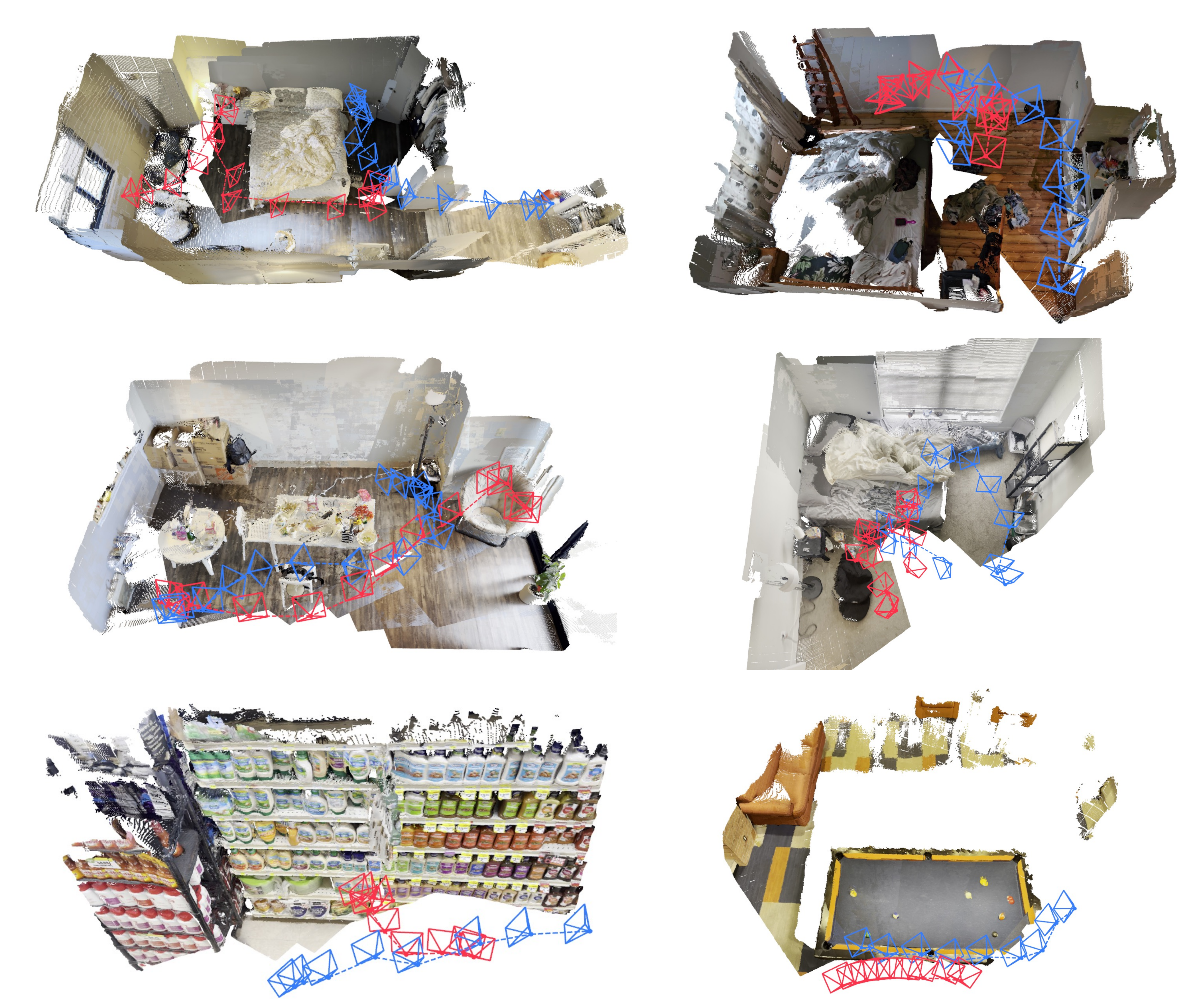}
\end{tabular}
\vspace{-1em}
\caption{\textbf{Camera Trajectory Visualization on SceneDiff Benchmark.} 
We visualize camera trajectories overlaid on point clouds predicted by $\pi^3$. \textcolor[RGB]{0,153,255}{Blue} cameras represent the sequence before the change, and \textcolor[RGB]{255,51,51}{red} cameras represent the sequence after the change.}
\label{fig:supp_camera_trajectories}
\vspace{-2em}
\end{figure}

We visualize camera trajectories for several scenes from the SceneDiff benchmark in Fig.~\ref{fig:supp_camera_trajectories}, demonstrating that 
our trajectories differ between the before and after sequences in the sequence pairs.

\subsection{RC3D Evaluation}
\label{supp:metric_rc3d}
In RC3D~\cite{Sachdeva2023TheCY}, methods are required to predict bounding boxes in both present and absent views, i.e., for one removed object, the bounding box of the object in the image captured after the change is also treated as a ground truth and evaluated. Therefore, to incorporate VLM models that cannot predict bounding boxes in absent views, we evaluate the bounding boxes from present views and absent views separately.
For our method, we predict the mask in the present view, unproject the pixels within the mask into 3D, and project to the absent view to retrieve the corresponding bounding box.

\subsection{Details about 3DGS-CD and MV3DCD}
\label{supp:3DGS-CD}

To better evaluate 3DGS-CD~\cite{Lu20243DGSCD3G} and MV3DCD~\cite{Galappaththige2024MultiViewPC} on the SceneDiff Benchmark, we sample 3 frames per second (rather than 1 FPS) when training the 3D Gaussian Splats~\cite{Kerbl20233DGS}, while still evaluating only on the 1 FPS sampled frames. We initially attempted to generate camera poses following the 3DGS-CD approach (COLMAP~\cite{colmapsfm} for pre-change poses and localization~\cite{Sarlin2018FromCT} with SfM point clouds for post-change poses). However, we found that our regressed poses were significantly more accurate, so we use our regressed camera parameters for both methods.
Since we only evaluate changed objects in visible views, we train 3DGS twice: once on the pre-change image sequence to render images from post-change poses and predict masks for \texttt{Added} or \texttt{Moved} objects, and once on the post-change image sequence to render images from pre-change poses and predict masks for \texttt{Removed} or \texttt{Moved} objects.

\begin{figure}[h]
\centering
\begin{tabular}{c}
        \includegraphics[width=0.8\linewidth]{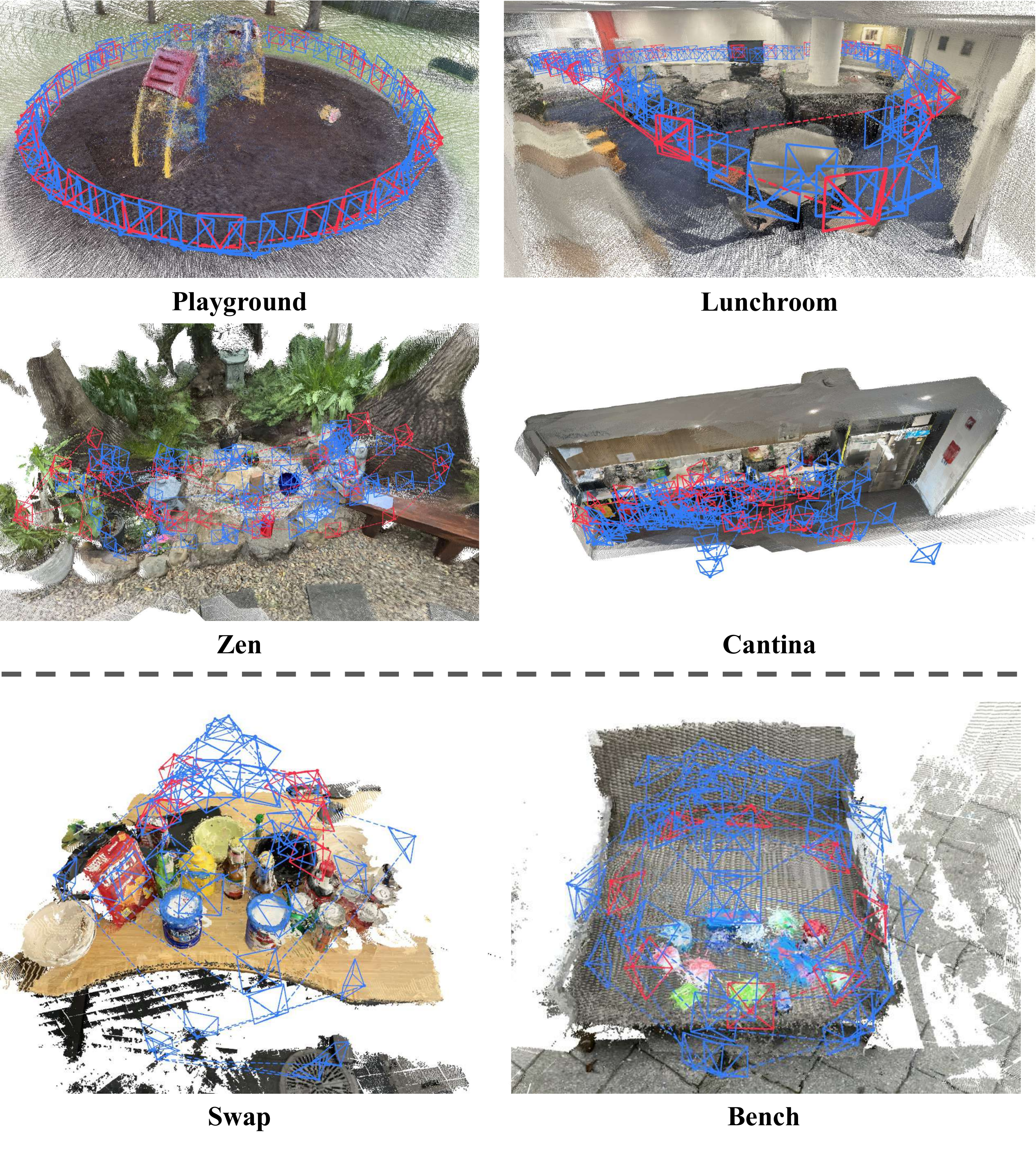}
\end{tabular}
\vspace{-1.5em}
\caption{
\textbf{Trajectory visualization of PASLCD and 3DGS-CD.} We visualize the pre-change (training) and post-change (test) camera trajectories. \textcolor[RGB]{0,153,255}{Blue} cameras represent the pre-change sequence, and \textcolor[RGB]{255,51,51}{red} cameras represent the post-change sequence. The pre-change sequence is densely captured, and the post-change poses are largely interpolated from the pre-change poses. The top panel shows PASLCD, and the bottom panel shows 3DGS-CD.
}
\label{fig:supp_3dgs_dataset_vis}
\end{figure}

\begin{figure}[h]
\hspace{-4em}
\centering
\begin{tabular}{c}
        \includegraphics[width=0.97\linewidth]{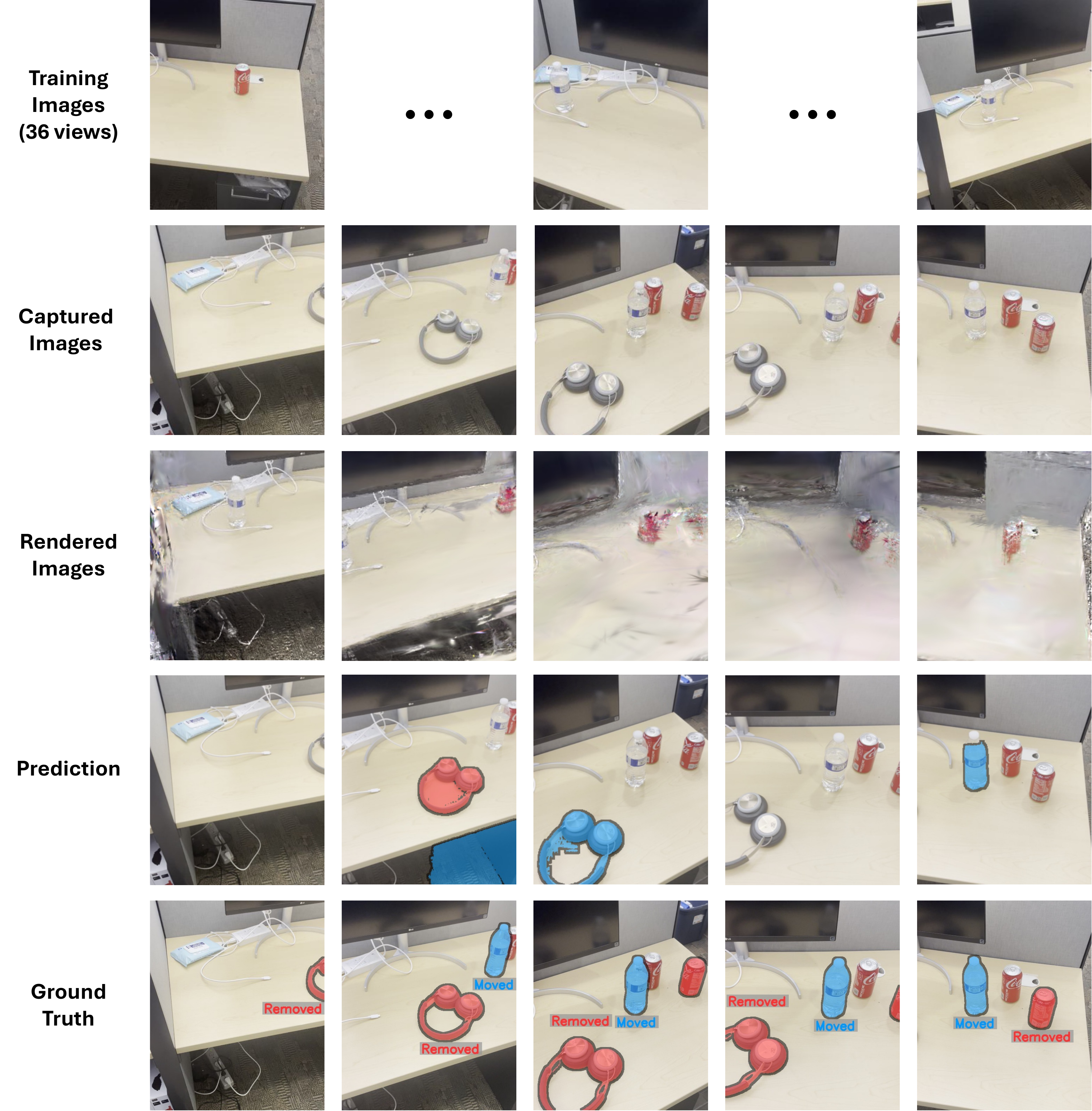}
\end{tabular}
\caption{
    \textbf{Visualization of 3DGS-CD.} The first row shows the images used to train the 3D Gaussian Splats before the change. The second and third row show the captured images after the change and rendered images of the pre-change 3DGS. The last two rows show the prediction and ground truth. We can see that the pose of rendered images are pretty accurate, but the rendered images are noisy, and therefore the method fails to detect many changed objects. Color map:
    \textcolor[RGB]{255,51,51}{\texttt{Removed}} and \textcolor[RGB]{0,153,255}{\texttt{Moved}}.
}
\label{fig:supp_3dgs_cd}
\end{figure}

We attribute the performance drop of 3DGS-based methods on SceneDiff benchmark primarily to the challenges of novel view synthesis under large viewpoint changes. As shown in Fig.~\ref{fig:supp_3dgs_dataset_vis}, the datasets originally used to evaluate methods like PASLCD and 3DGS-CD feature densely captured scenes. In these settings, the post-change (test) poses are largely covered by the pre-change (training) sequences, allowing the models to rely on view interpolation to produce high-quality renders.
In contrast, the pre-change and post-change trajectories in the SceneDiff benchmark differ significantly (Fig.~\ref{fig:supp_camera_trajectories}). This introduces the more difficult task of view extrapolation from sparse inputs. Because 3DGS-based change detection depends heavily on accurate rendering, these larger viewpoint shifts can introduce rendering artifacts (illustrated in Fig.~\ref{fig:supp_3dgs_cd}), which correspondingly affect the downstream change detection accuracy.

For 3DGS-CD, we follow their codebase and use the SAM prediction confidence as the detection confidence. They also build a 3D occupancy grid of all changed regions from input views and project the grid back to all views. However, we found that the occupancy grid is inaccurate due to poor underlying geometry, causing a significant drop in \textit{obj/im AP}. Therefore, we do not use the occupancy grid in our evaluation.

For MV3DCD, we render the symmetric change predictions and follow the 3DGS-CD approach, utilizing SAM to assign changes to the correct sequence (i.e., determining whether a change belongs to the pre-change or post-change sequence). However, because MV3DCD does not predict instance-level changes like 3DGS-CD, we must adapt our use of SAM. Specifically, we extract non-overlapping regions in the current view and compute the ratio of changed pixels within each region. A region is classified as positive if this ratio exceeds an empirically chosen 60\% threshold.
Pixels within these positive regions are then assigned to the current view, which filters out noise and changes visible only in the other sequence. Recognizing that this adaptation might introduce a slight disadvantage for MV3DCD, we also evaluate our method on their proposed dataset to ensure a comprehensive comparison (Sec.~\ref{supp:PASLCD}), where our method still achieves slightly better results in the densely captured scenes.

\clearpage
\subsection{Details about VLM}
\label{supp:VLM}
We present the text prompts used for two-view and two-sequence change detection tasks. The text outputs are then fed into SAM3~\cite{Carion2025SAM3S} to localize the changed objects.
For detection scoring, we use the prediction confidence from SAM3 rather than the VLM because we empirically found that VLMs tend to output discrete, uncalibrated confidence values (e.g., 0.3, 0.5, 0.7) and fail to provide reliable, continuous confidence scores across diverse scenes.
\paragraph{Two-view text prompt input:}
\footnotesize
{\begin{verbatim}
You are a helpful computer vision assistant. The two input images are 
stitched together side by side, left is the first one and right is the 
second one. The two input images show the same scene and might be 
captured from different viewpoints.

Please list all objects and their positions in the scene for each image 
independently. Next, Compare the two images carefully and identify any 
object-level changes even if they are subtle. There are three types of 
changes:

1. added: The object appears in the second image but not in the first one.
2. removed: The object appears in the first image but not in the second 
one.
3. moved: The object appears in both images but its position has changed.

Return the list in structured JSON format, e.g., 
[{"object": "bottle", "change": "removed"}, ...]

First Image: image_1
Second Image: image_2

\end{verbatim}}
\paragraph{Two-sequence text prompt input:}
\footnotesize{
\begin{verbatim}
You are a helpful computer vision assistant. The two input videos show 
the same scene captured at different time. Please list all objects and
their positions in the scene for each video independently.

Next, Compare the two videos carefully and identify any object-level 
changes even if they are subtle, ignoring the effect of viewpoint change. 
There are three types of change:

1. added: The object appears in the second video but not in the first one.
2. removed: The object appears in the first video but not in the second 
one.
3. moved: The object appears in both videos but its position has changed.

Return the list in structured JSON format, e.g., 
[{"object": "bottle", "change": "removed"}, 
{"object": "ball", "change": "added"}, ...]

First video: video_1_frames
Second video: video_2_frames
\end{verbatim}}

\subsection{Details about CYWS-2D}
\label{supp:CYWS-2D}
To comprehensively evaluate CYWS-2D, we curate 3,125 paired frames from our validation set. First, we sample frames from corresponding sequences at 1 FPS. For each view in a sequence, we identify views in the paired sequence that share over 50\% co-visibility. For every valid pair, we extract the bounding boxes of changed objects in the source view, strictly ensuring these objects are also visible in the target view. This process is applied across the entire validation set to construct the final evaluation pairs (paired examples are shown below). Finally, because CYWS-2D is already fine-tuned on directional change data, it does not require any post-processing during evaluation. 

For our primary evaluation, the predicted bounding boxes are converted into segmentation masks using SAM. To ensure this conversion step does not negatively impact the results, we also evaluate the direct bounding box predictions using an IoU threshold of 0.5. The direct box evaluation yields an \textit{obj/im AP} of 21.9 on SD-V and 14.5 on SD-K. When evaluated using the SAM-generated masks, the performance actually increases to 24.5 and 16.8, respectively. This confirms that relying on SAM for mask generation does not cause a performance drop, but rather effectively translates the bounding boxes into accurate evaluation masks.
\begin{figure}[h]
\centering
\vspace{-1em}
\begin{tabular}{c}
        \includegraphics[width=0.98\linewidth]{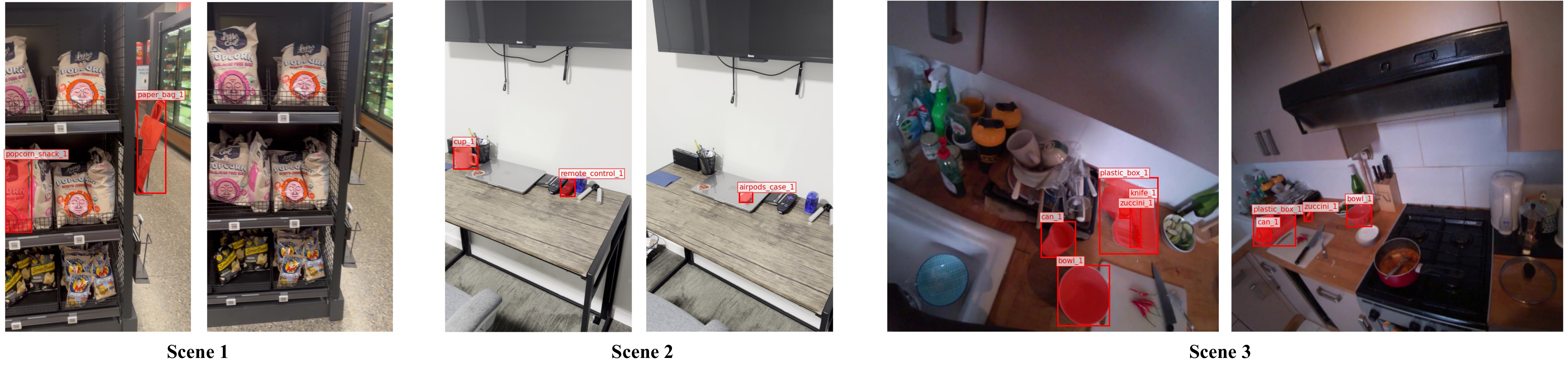}
\end{tabular}
\vspace{-1em}
\caption{
\textbf{Extracted Training Examples for CYWS-2D.} The first two scenes are from SD-V, and the third is from SD-K.
}
\vspace{-1.5em}
\end{figure}

\clearpage
\subsection{Details about CYWS-3D}
\label{supp:CYWS-3D}
CYWS-3D~\cite{Sachdeva2023TheCY} generates paired bounding box predictions across two input views. To adapt this for directional change evaluation on the SceneDiff benchmark, we also consider the prediction in the paired view. While true positives are converted to masks and evaluated normally, we apply a specific filtering rule for false positives: if a predicted box in the source view is a false positive, but its corresponding pair in the alternate view is a true positive, the initial false positive is discarded. Finally, to ensure a fair comparison, we provide CYWS-3D with depth maps regressed by $\pi^3$~\cite{wang2025pi3} during inference.

Similar to our evaluation of CYWS-2D, the predicted bounding boxes from CYWS-3D are To ensure that this mask generation step does not unfairly degrade the model's scores, we provide a secondary evaluation using the raw bounding boxes directly (at an IoU threshold of 0.5). Under this box-level metric, CYWS-3D achieves an \textit{obj/im AP} of 10.1 on SD-V and 4.3 on SD-K. By contrast, evaluating with the SAM-extracted masks boosts these metrics to 13.9 and 6.6, respectively. This demonstrates that our SAM-based adaptation does not hinder the baseline, but rather serves as an effective bridge from box-level to mask-level evaluation.

\begin{figure}[h]
\vspace{-1em}
\centering
\begin{tabular}{c}
        \includegraphics[width=0.95\linewidth]{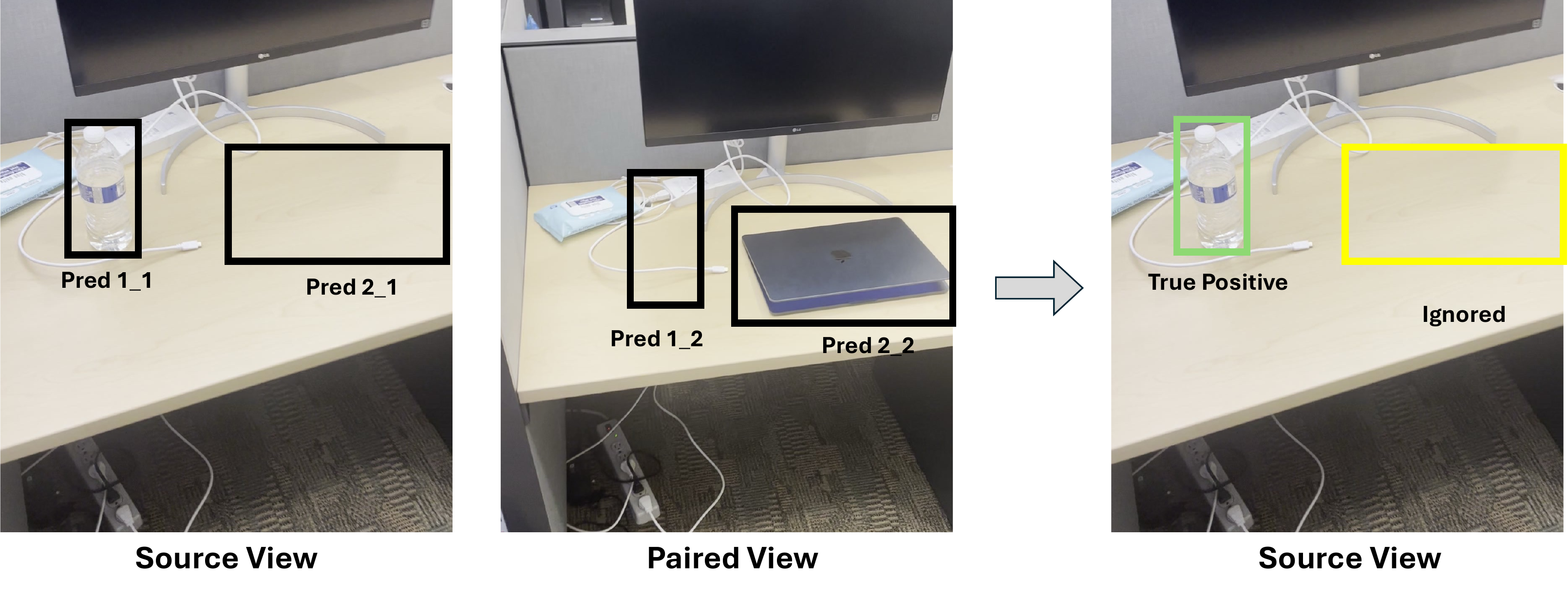}
\end{tabular}
\vspace{-1.2em}


    \caption{
\textbf{CYWS-3D Evaluation on the SceneDiff Benchmark.}
In this example, CYWS-3D predicts two box pairs: (Pred 1\_1, Pred 1\_2) and (Pred 2\_1, Pred 2\_2). During source view evaluation, we first check if a prediction is a \textcolor[RGB]{51, 204, 102}{\texttt{True Positive}}. If not (e.g., Pred 2\_1), we check whether its paired box in the alternate view (Pred 2\_2) matches any ground truth. If the paired box is a match, the source prediction (Pred 2\_1) is \textcolor[RGB]{255,233,0}{\texttt{Ignored}}; otherwise, it remains a \textcolor[RGB]{255,51,51}{\texttt{False Positive}}.
}
\label{fig:supp_cyws_3d_vis}
\vspace{-2em}
\end{figure}

\clearpage
\section{SceneDiff Benchmark}
\label{supp:scenediff}

\subsection{Sequence Pair Annotation Example}
\label{supp:annotation}
\begin{figure}[h]
\centering
\begin{tabular}{c}
        \includegraphics[width=0.67\linewidth]{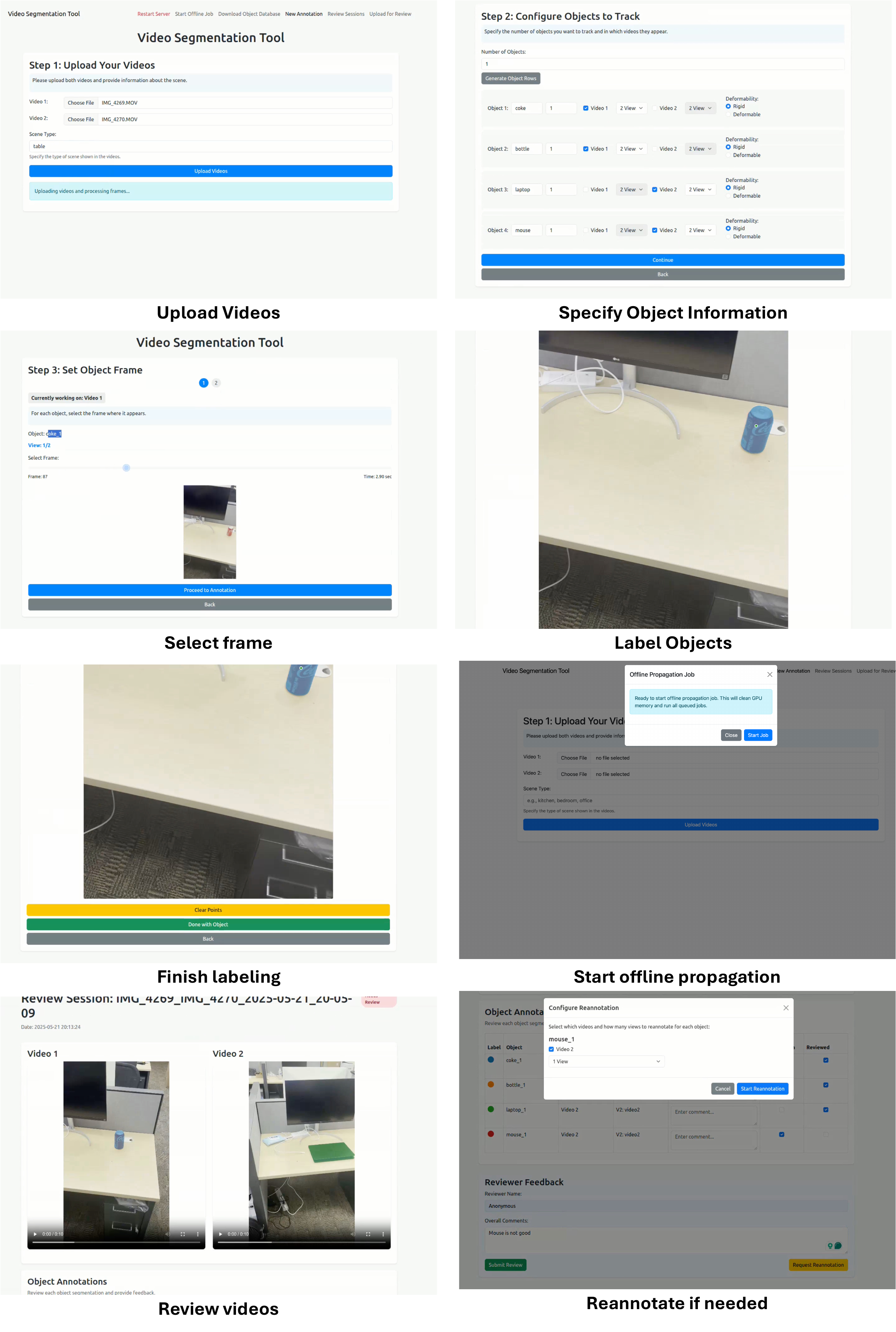}
\end{tabular}

\caption{
\raggedright
    \textbf{Visualization of Sequence Pairs Annotation.} 
}
\label{fig:supp_annotation_vis}
\end{figure}

We visualize the key steps of annotating sequence pairs in Fig.~\ref{fig:supp_annotation_vis}, but we recommend viewing the annotation video attached inside the zip file (we skip the offline propagation waiting time in the video, which typically takes around 5 minutes for one sequence pair).
The process consists of the following steps: (1) upload video sequences, (2) fill in object information (automatically populated from Google Sheets if already filled in), (3) select key frames, (4) label objects, (5) complete manual labeling, (6) start offline propagation, (7) review annotated videos, and (8) reannotate objects or submit annotated videos.

\subsection{Data Collection Instruction}
\label{supp:collection}
\includegraphics[width=0.95\textwidth,page=1]{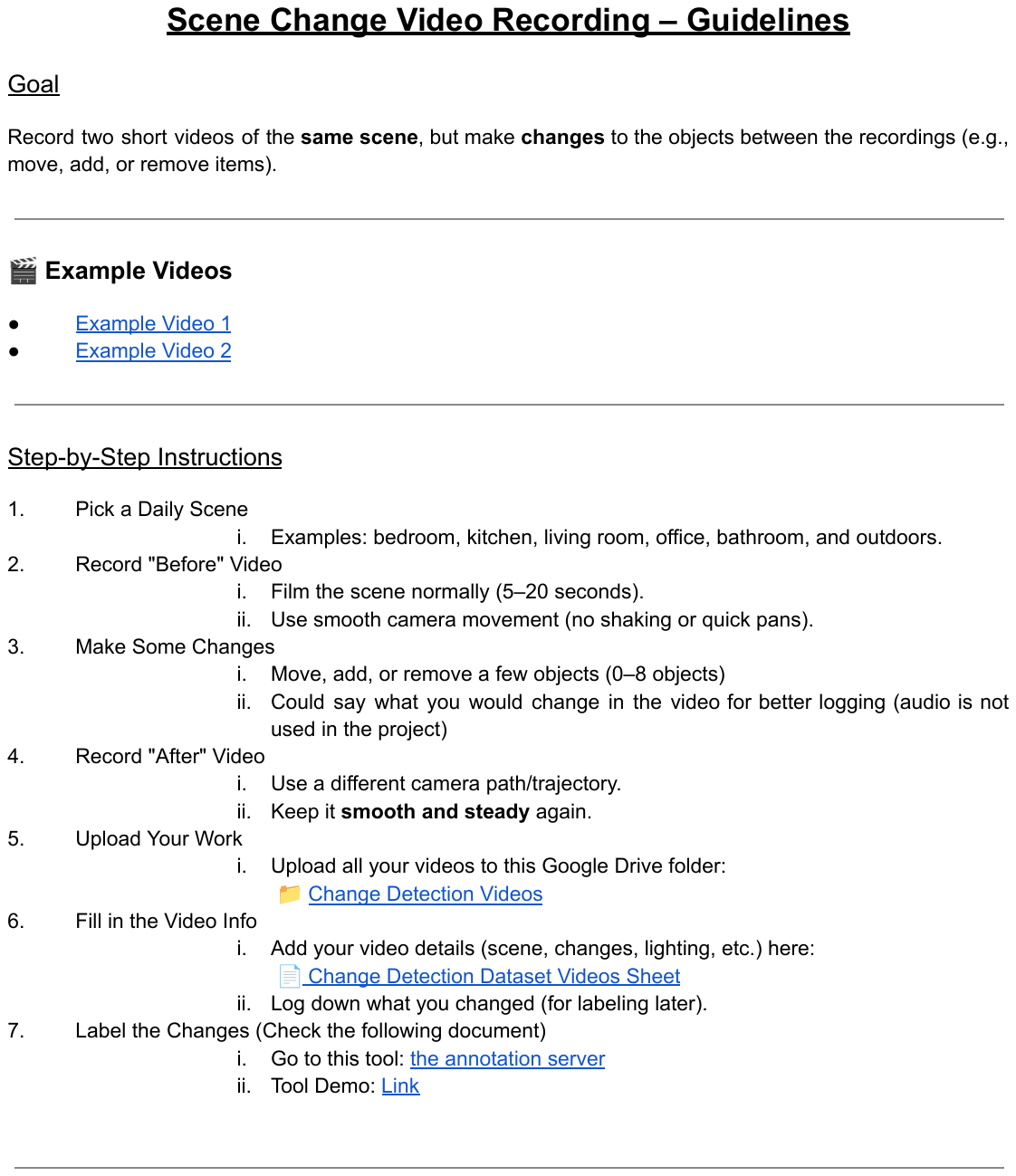}
\clearpage
\foreach \n in {2}{  
    \begin{center}
        \includegraphics[page=\n,width=0.95\textwidth,keepaspectratio]{images/scene_change_detection_guidelines_for_supp.docx.pdf}
    \end{center}
    \clearpage
}


\end{document}